\documentclass[11pt]{article}
\usepackage[final]{acl}
\usepackage{float}
\usepackage[utf8]{inputenc}
\usepackage{CJKutf8}
\usepackage{times}
\usepackage{CJKutf8}
\usepackage{tabularx}
\usepackage{booktabs}
\usepackage{latexsym}
\usepackage{booktabs}     
\usepackage{multirow}     
\usepackage{threeparttable} 
\usepackage{amsmath}
\usepackage{enumitem}
\usepackage{amssymb}
\usepackage{threeparttable}
\usepackage[T1]{fontenc}
\usepackage{float}  
\usepackage{subcaption}  

\usepackage[utf8]{inputenc}

\usepackage{microtype}
\usepackage{graphicx}

\usepackage{inconsolata}

\usepackage{graphicx}

\title{On Temperature-Constrained Non-Deterministic Machine Translation: Potential and Evaluation}

\author{
    Weichuan Wang\textsuperscript{1,2}, 
    Mingyang Liu\textsuperscript{1}, 
    Linqi Song\textsuperscript{1,2}\thanks{Corresponding authors.}, 
    Chen Ma\textsuperscript{1,2}\footnotemark[1] \\
    \textsuperscript{1}City University of Hong Kong \\
    \textsuperscript{2}City University of Hong Kong Shenzhen Research Institute\\
    \texttt{\{weicwang2-c, mingyaliu8-c\}@my.cityu.edu.hk} \\
    \texttt{\{linqsong, chenma\}@cityu.edu.hk}
}

\usepackage{fancyhdr}
\begin{document}
\maketitle
\thispagestyle{fancy}

\begin{abstract}
In recent years, the non-deterministic properties of language models have garnered considerable attention and have shown a significant influence on real-world applications. However, such properties remain under-explored in machine translation (MT), a complex, non-deterministic NLP task. In this study, we systematically evaluate modern MT systems and identify temperature-constrained \textbf{N}on-\textbf{D}eterministic \textbf{MT} (\textbf{ND-MT}) as a distinct phenomenon. Additionally, we demonstrate that ND-MT exhibits significant potential in addressing the multimodality issue that has long challenged MT research and provides higher-quality candidates than \textbf{D}eterministic MT (D-MT) under temperature constraints. However, ND-MT introduces new challenges in evaluating system performance. Specifically, the evaluation framework designed for D-MT fails to yield consistent evaluation results when applied to ND-MT. We further investigate this emerging challenge by evaluating state-of-the-art ND-MT systems using both lexical-based and semantic-based metrics at varying sampling sizes. The results reveal a Buckets Effect across these systems: the ranking of ND-MT systems is dominated by the worst-quality candidate translation, as shown by automatic evaluation metrics. To mitigate this issue, we propose ExpectoSample, a strategy that first identifies reliable metrics and then enables robust ND-MT system selection for real-world.

\end{abstract}

\section{Introduction}

The revolutionary development of large language models and their emergent capabilities~\cite{DBLP:journals/tmlr/WeiTBRZBYBZMCHVLDF22} has demonstrated significant influence across various fields, including complex downstream NLP tasks~\cite{DBLP:conf/nips/WangPNSMHLB19,DBLP:conf/iclr/HendrycksBBZMSS21,li-etal-2024-cmmlu}, science~\cite{dsouza-etal-2025-yescieval}, and mathematical reasoning~\cite{DBLP:conf/eacl/AhnVLLZY24}. In recent years, researchers have increasingly recognized the benefits of the non-deterministic (ND) properties~\cite{atil2025nondeterminismdeterministicllmsettings, song-etal-2025-good} of LLMs on their potential for flexible and customized chat-box applications~\cite{deepseek,openai2024gpt4technicalreport,qwen3}. Recent studies have progressed fine-grained exploration and analysis of this property, primarily focusing on deterministic tasks such as question answering~\cite{song-etal-2025-good, smantic_ent}. However, the impact of ND on machine translation, a complex and ND task, remains under-explored both on the generation and evaluation aspects. To bridge this research gap, we examine modern \textbf{N}on-\textbf{D}eterministic \textbf{Machine Translation} (\textbf{ND-MT}) systems for both the potential benefits and the evaluation challenges associated with non-determinism.

We first examine one of the most prominent challenges in MT: multimodality \cite{bleu, bao-etal-2023-non}, a phenomenon where a single source sentence may correspond to multiple valid translation candidates due to contextual ambiguity. This challenge is particularly problematic for automatic evaluation due to the scarcity of comprehensive reference sets \cite{bleu, chrf, rei-etal-2022-cometkiwi}. While previous research has employed human assessment \cite{kocmi-etal-2024-findings, kocmi-etal-2023-findings, kocmi-etal-2025-findings} to mitigate this issue, such approaches are increasingly unscalable due to the prohibitive cost of constant re-evaluation necessitated by domain shifts in source texts \cite{kocmi-etal-2025-findings}. 
We reformulate this challenge as a dual requirement for MT: the candidates for a source sentence should demonstrate lexical diversity~\cite{ploeger-etal-2024-towards} while maintaining semantic equivalence~\cite{smantic_ent} with the original source sentence. Notably, candidates generated from ND-MT have the potential to satisfy both principles simultaneously. This is observed in practical applications where the same translation prompt yields diverse yet acceptable outputs. 
In this work, we systematically investigate the potential of ND-MT for addressing multimodality, focusing on its ability to provide both lexical diversity and semantic equivalence. Our analysis covers 22 modern MT systems across six language directions under a uniform temperature setting ($0.5$) (see Appendix~\ref{model_statistics} for details). Additionally, we introduce a reference-free, group-level lexical metric termed the Group Lexical Variance Score (GLVS) to mitigate the bias resulting from limited references when evaluating ND-MT systems. We employ both lexical and semantic metrics to quantify the impact of non-determinism on lexical variability and meaning preservation, respectively. Our results demonstrate significant lexical diversity alongside nearly identical semantic content when compared to D-MT systems using the same underlying models. 
Furthermore, we investigate how the value of temperature affects the final system performance of ND-MT. The results reveal that while various temperature settings can induce lexical diversity, semantic equivalence is only preserved at lower temperatures. Consequently, we characterize modern MT systems as temperature-constrained ND-MT systems.


However, the evaluation of ND-MT remains under-explored. We first investigate the feasibility of directly adopting performance from D-MT counterparts using current evaluation schemes \cite{kocmi-etal-2024-findings, kocmi-etal-2023-findings, kocmi-etal-2025-findings}. To bypass the high cost of human assessment, we explore this direct adaptation through existing automatic evaluation metrics \cite{bleu, chrf, rei-etal-2022-cometkiwi} that exhibit high correlation with human judgment. Specifically, we apply these via group-level measurements: \emph{min}, \emph{max}, \emph{mean}, \emph{random}, and \emph{std} (standard deviation), capturing the group-level performance of ND-MT systems. The resulting inconsistent rankings demonstrate the unreliability of traditional D-MT evaluation schemes when applied to the ND-MT. Furthermore, we examine how sampling size ($\{10, 20, 50\}$) affects evaluation results with five state-of-the-art ND-MT systems at a fixed temperature ($0.5$). The results reveal a strong Buckets Effect: for each source, the lowest-quality candidate largely determines system rankings across sample sizes. This highlights the inherent risk in evaluating non-deterministic systems, as the minimum performance quality is stochastically hidden and cannot be predicted before generation. To mitigate this, we propose the ExpectoSample strategy, which first identifies reliable metrics and subsequently selects robust ND-MT systems.


Our contributions are threefold: (1) We demonstrate that ND-MT systems effectively address the multimodality challenge by providing lexical diversity while maintaining semantic equivalence under specific temperature constraints. (2) We uncover the Buckets Effect in ND-MT evaluation—where system rankings are predominantly determined by the lowest-quality candidates, and propose the ExpectoSample strategy to mitigate this challenge by identifying reliable metrics for robust system selection. (3) We conduct a systematic investigation of 22 ND-MT systems across six language directions using 11,947 source cases. To support future research, we release our complete code, dataset, and evaluation results at \footnote{https://github.com/weichuanW/TC-DN-MT}.

\section{Related Works}
\subsection{Modern MT Systems}
Modern machine translation follows the sequence-to-sequence paradigm~\cite{DBLP:conf/nips/SutskeverVL14} with the Transformer~\cite{DBLP:journals/corr/VaswaniSPUJGKP17} as the backbone and is divided into two main types: encoder-decoder models pre-trained~\cite{DBLP:journals/corr/VaswaniSPUJGKP17} on multilingual text then fine-tuned on bilingual text~\cite{nllb-200}, and decoder-only architectures pre-trained on multilingual text without specific fine-tuning requirements~\cite{brown2020language}. From the inference perspective, encoder-decoder models~\cite{mbart-50,nllb-200} require explicit language signals as input during both training and inference, while decoder-only models~\cite{llama2,llama3,qwen2.5,qwen3,deepseek} leverage the inherent multilingual semantic alignment of LLMs and activate MT capabilities through prompts~\cite{vilar-etal-2023-prompting}. Different LLM-based MT approaches exhibit distinct characteristics: pre-training-only MT systems typically use few-shot methods~\cite{DBLP:journals/corr/abs-2005-14165,vilar-etal-2023-prompting} (commonly five-shot) but inevitably introduce repetition and language mismatch issues~\cite{wang-etal-2024-mitigating-language}; instruction-tuned MT systems use direct MT prompts but sometimes produce noise without strict constraints~\cite{llama2, llama3} (e.g., Chinese translations including Pinyin in Llama series models); RL-based reasoning MT systems use direct MT prompts and can provide detailed translation steps but require substantial computational resources for both post-editing and inference~\cite{deepseek,qwen3}. Generally, modern MT systems use a generate-once approach~\cite{kocmi-etal-2025-findings} to produce deterministic results, while their potential to generate multiple candidate translations through non-deterministic sampling remains under-explored.
\subsection{Non-determinism of LLMs}
Previously, substantial effort was focused on deterministic tasks such as sentiment classification \cite{zhang-etal-2024-sentiment} and parsing \cite{ginn-palmer-2025-llm}, with most attention directed toward utilizing deterministic capabilities from LLMs. In recent years, the non-determinism (ND) property of LLMs has emerged as a significant area of interest and have been leveraged to satisfy customized user requirements \cite{tseng-etal-2024-two}. Most models now implement non-determinism as a default property \cite{deepseek, qwen3}, enabling LLMs to provide a variety of reasonable outputs for the same prompt to enhance user satisfaction. Previous studies have found that this property can benefit certain deterministic NLP tasks \cite{song-etal-2025-good}, such as question answering, by generating semantically equivalent responses \cite{smantic_ent}. However, research of ND of LLMs in complex tasks like MT remains underexplored.
\subsection{Automatic Evaluation on MT}
Automatic evaluation methods play a key role in evaluating MT systems by avoiding the substantial costs of human assessment. In this work, we investigate the potential of ND-MT to provide lexical diversity and semantic equivalence. To achieve this goal, we categorize current metrics into two main categories: lexical-based methods and semantic-based methods, to measure the capabilities of ND-MT.
For lexical-based methods, BLEU~\cite{bleu}, METEOR~\cite{meteor}, and ROUGE~\cite{rouge}, which focus on lexical overlap. ChrF++~\cite{chrf} focuses on character overlap and TER~\cite{ter} focuses on error edit distance. Specifically, these methods rely on references, suffer from the multimodality issue, and fail without references.
For semantic-based methods, BERTScore~\cite{bertscore} and BLEURT~\cite{bleurt} utilize the token information to model the semantic score. COMET20-DA~\cite{comet20da} and COMET22-KIWI~\cite{comet22kiwi} include a training stage to learn the semantic equivalence between source and candidates. XCOMET~\cite{xcomet} further evaluates on the error spans. Other methods measure semantic alignment through semantic similarity between the source and candidates in a unified semantic embedding space. including SentTrans~\cite{sentrans} with direct LMs, LASER~\cite{laser}, and XNLI~\cite {xnli} using bilingual pairs.

\section{Modern MT Systems Are Temperature-Constraint ND-MT}\label{sec_temp_constraint}
In this section, we systematically investigate the non-deterministic properties of modern MT systems. We begin by selecting state-of-the-art architectures, including both encoder-decoder and decoder-only models of varying scales. We then generate multiple translation candidates and evaluate them using automatic metrics via group-level measurements. Our analysis reveals that ND-MT effectively addresses the multi-modality challenge by providing lexical diversity while maintaining semantic equivalence under specific temperature constraints. Based on these observations, we characterize modern MT systems as temperature-constrained ND-MT systems.

\subsection{Experimental Preparation}
\subsubsection{ND-MT Systems}
We consider both encoder-decoder and decoder-only architectures for modern MT. For encoder-decoder architectures, we select mBART~\cite{mbart-50} trained on 50 multilingual texts (0.68B parameters) and NLLB-200~\cite{nllb-200} with three model scales (0.6B, 3.3B, and 54.6B parameters). For LLM-based MT, we include the Llama-2 series~\cite{llama2}, Llama-3 series~\cite{llama3}, Qwen-2.5 series~\cite{qwen2.5}, Qwen-3~\cite{qwen3} series, and DeepSeek series~\cite{deepseek}, examining both small-scale (7 and 8B parameters) and large-scale (70, 72 and 671B parameters) variants across pre-trained, instruction-tuned, and reasoning types when available. The detailed information are provided in Appendix~\ref{model_statistics}.
\subsubsection{Dataset Statistics}
\begin{table}[t]
\centering
\caption{Dataset Statistics Information}
\label{tab:dataset_statistics}
\small
\begin{tabular}{@{}llr@{}}
\toprule
Source & Translation Direction & Size \\
\midrule
WMT23 & En$\rightarrow$Zh & 2,074 \\
WMT23 & Zh$\rightarrow$En & 1,976 \\
WMT23 & En$\rightarrow$De & 557 \\
WMT23 & De$\rightarrow$En & 549 \\
WMT23 & En$\rightarrow$Ru & 2,074 \\
WMT23 & Ru$\rightarrow$En & 1,723 \\
WMT24 & En$\rightarrow$Zh & 998 \\
WMT24 & En$\rightarrow$De & 998 \\
WMT24 & En$\rightarrow$Ru & 998 \\
\bottomrule
\end{tabular}
\end{table}
In this work, we adopt sentence-level MT and leverage existing, well-established open-source datasets to study both ND-MT and their counterpart D-MT systems. Specifically, we use the latest WMT data from 2023–2024~\footnote{https://github.com/wmt-conference/wmtX-news-systems, x = $\{23, 24\}$} across six translation directions (ZH$\leftrightarrow$EN, EN$\leftrightarrow$DE, EN$\leftrightarrow$RU), covering three language pairs. We identify ⟨English, Chinese⟩ translation as particularly valuable for investigation due to substantial differences in language families and choose it as our primary experimental setting to explore the potential of ND-MT. Further evaluation on ⟨English, German⟩ and ⟨English, Russian⟩ are made to demonstrate the general potential of ND-MT across diverse language pairs. We present detailed statistics in Table \ref{tab:dataset_statistics}.

\subsubsection{Evaluation Methods}\label{eval_methods}
\paragraph{Lexical-based Methods}
We include BLEU~\cite{bleu}, an n-gram-based metric evaluating lexical overlap; ChrF++~\cite{chrf}, an n-gram-based metric capturing both lexical and character-level information; METEOR~\cite{meteor}, a token-level alignment metric; ROUGE(-1, -2, -L)\cite{rouge}, a recall-oriented n-gram overlap metric; and TER~\cite{ter}, a token-level edit distance metric. These metrics use the reference as an anchor to show the lexical diversity.

\paragraph{Semantic-based Methods}
For semantic equivalence, we employ COMETKIWI\cite{comet22kiwi} and COMETDA~\cite{comet20da} to measure with the neural network; LASER~\cite{laser}, LaBSE~\cite{laser}, SentTrans~\cite{sentrans}, and XNLI~\cite{xnli} to test the semantic equivalence on a unified semantic space; BLEURT~\cite{bleurt} and BERTScore~\cite{bertscore} to measure the semantic equivalence with token information.

\paragraph{Group Lexical Variance Score (GLVS)}
A primary drawback of current lexical-based metrics is their reliance on gold-standard references~\cite{bleu,chrf,rouge,ter}, which are typically unavailable for ND-MT systems in real-world scenarios. While some in-group lexical evaluation methods~\cite{zhu2018texygen} adapt the principles of BLEU~\cite{bleu} or ChrF++~\cite{chrf} to measure overlapping between candidates, these approaches often lack discriminative power. Specifically, when candidates are highly similar or significantly different, these metrics tend toward extreme values (zero or one), failing to capture nuanced lexical diversity. Conversely, simple word-counting strategies~\cite{liu2022rethinking} mitigate these extreme cases but fail to account for the lexical relationships, such as shared vocabulary between candidates. To address these limitations, we propose the Group Lexical Variance Score (GLVS) to quantify lexical diversity by establishing a group-level vocabulary and computing the frequency distribution of unique tokens across the candidate set, capturing both individual variance and inter-candidate relationships.

The computation of the Group Lexical Variance Score (GLVS) proceeds in three stages:

\textbf{1. Candidate Tokenization}
Each candidate $c_i$ in the generated set $\mathcal{C} = \{c_1, c_2, \dots, c_N\}$ is tokenized into a sequence of words $\mathcal{W}_i = \{w_1, w_2, \dots, w_l\}$. To focus on lexical variety, we define $\mathcal{W}_i^U$ as the set of unique words for candidate $c_i$.

\textbf{2. Collective Vocabulary Frequency}
We define the collective word pool $\mathcal{V}_{total}$ as the vocabulary containing all words across all $N$ candidates. Let $M$ be the total word count in $\mathcal{V}_{total}$. For each unique word $w$, its relative frequency $f(w)$ is calculated as:\begin{equation}f(w) = \frac{\text{count}(w, \mathcal{V}_{total})}{M}\end{equation}

\textbf{3. GLVS Computation and Aggregation}
For each individual candidate $c_i$, we calculate a candidate-level score by summing the relative frequencies of its unique constituent words:\begin{equation}GLVS(c_i) = \sum_{w \in \mathcal{W}i^U} f(w)
\end{equation}
To characterize the overall behavior of an ND-MT system for a specific source sentence, we aggregate these individual scores at the group level using the mean ($\mu_{GLVS}$):
\begin{equation}
    \mu_{GLVS} = \frac{1}{N} \sum_{i=1}^{N} GLVS(c_i)
\end{equation}
In this framework, $\mu_{GLVS}$ serves as an inverse proxy for lexical diversity. A low $\mu_{GLVS}$ value indicates that candidates are primarily composed of words that appear rarely within the group, thereby signaling high lexical diversity. Conversely, a high value suggests lexical redundancy, where candidates converge on a narrow set of common terms.

Furthermore, other group-level measurements, like the standard deviation, can provide additional analytical depth.
For instance, a high standard deviation indicates that the ND-MT system is unstable in its generation quality, producing candidates with significantly varying degrees of lexical diversity.


\subsubsection{Experimental Settings}
\paragraph{Decoding Strategy}
We employ greedy decoding as the deterministic baseline and sampling-based decoding in the non-deterministic setting with adjustable temperature, generating $K$ candidates per source. We use temperature $0.5$ and sampling size $10$, motivated by a previous study~\cite{smantic_ent}, to investigate the potential of ND-MT.

\paragraph{Group-based Measurements}
For each source, we compute their group-based measurements: \emph{min}, \emph{max}, \emph{mean}, \emph{random}, and \emph{std} (standard deviation) for various metrics and aggregate them on the dataset level to capture the system performance of ND-MT.




\subsection{The Potential of ND-MT to Solve Multimodality}\label{potential_nd_mt}

\begin{figure*}[!t]
    \centering
    \begin{subfigure}{\textwidth}
        \centering
        \includegraphics[scale=0.5]{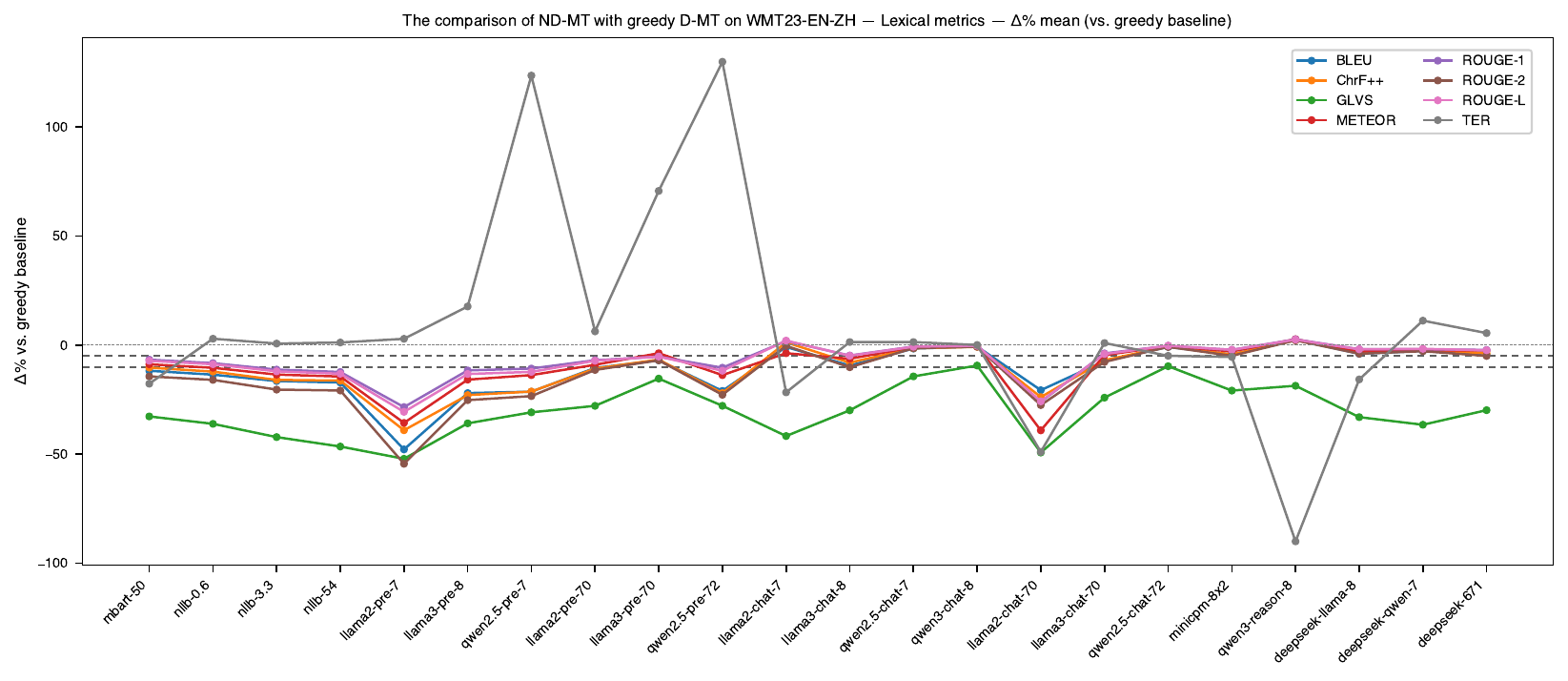}
        \caption{Mean Delta Results of WMT23 En$\rightarrow$Zh on Lexical Metrics.}
        \label{fig:delta_lexical_mean_en-zh}
    \end{subfigure}

    \vspace{0.5em}

    \begin{subfigure}{\textwidth}
        \centering
        \includegraphics[scale=0.5]{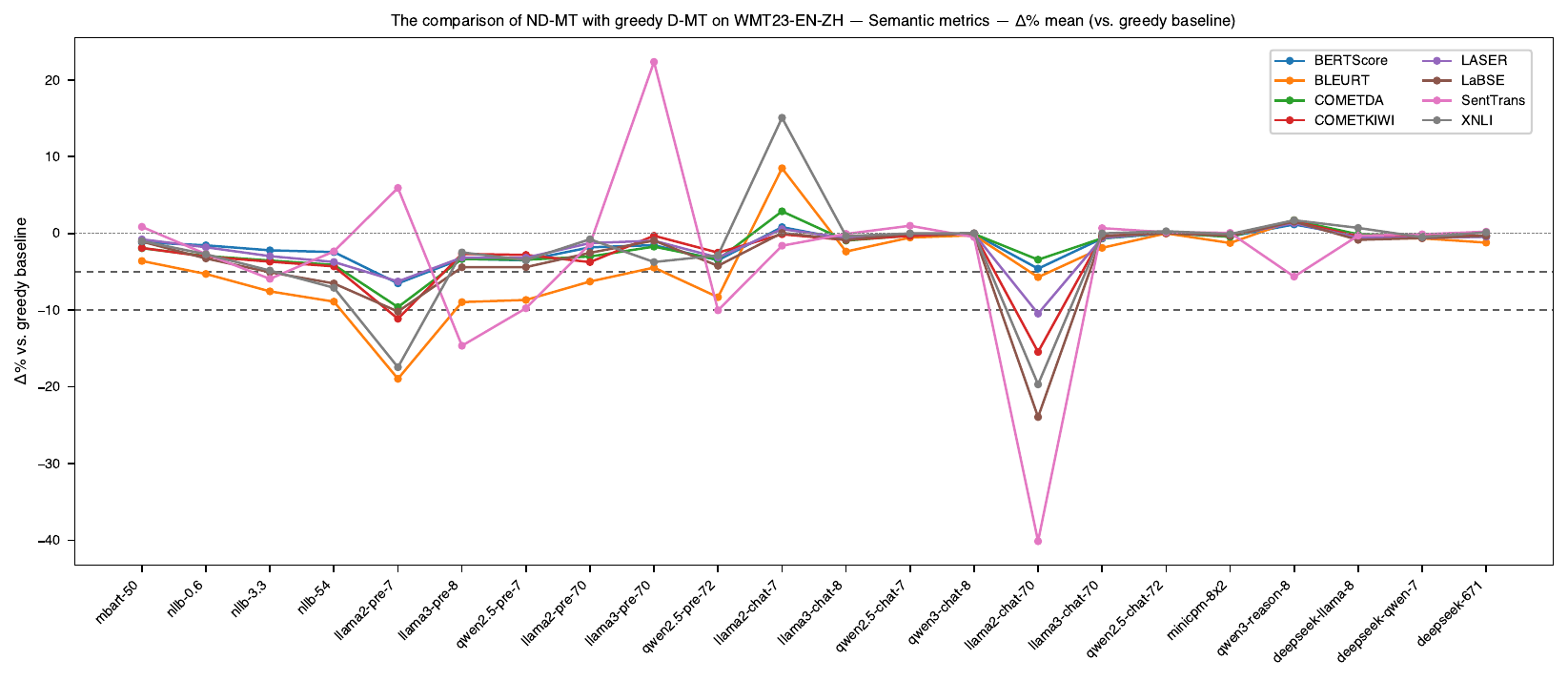}  
        \caption{Mean Delta Results of WMT23 En$\rightarrow$Zh on Semantic Metrics.}
        \label{fig:delta_semantic_mean_en-zh}
    \end{subfigure}

    \caption{Mean Delta results for lexical and semantic metrics on WMT23 En$\rightarrow$Zh ($T=0.5$, 10 candidates). Delta results are calculated relative to greedy decoding on identical data and models. Thresholds of $-5$ and $-10$ (dotted line) are included to indicate levels of significance.}
    \label{fig:delta_all_en-zh_mean}
\end{figure*}

To explore the potential of ND-MT in addressing the multimodality challenge across two dimensions: lexical variance and semantic equivalence. We conduct experiments on 22 ND-MT systems for the ⟨English, Chinese⟩ pair, with a temperature of $0.5$ and a sampling size of $10$~\cite{smantic_ent}, without additional non-deterministic settings such as top-p~\cite{DBLP:conf/iclr/HoltzmanBDFC20}, top-k~\cite{DBLP:journals/corr/abs-2505-19371}. We run the corresponding D-MT systems as baselines to enable direct comparison, where each system generates only one candidate.



\paragraph{ND-MT provides salient lexical diversity} Figure~\ref{fig:delta_lexical_mean_en-zh} illustrates the lexical diversity captured by various evaluation metrics. We first observe that traditional reference-based metrics~\cite{bleu,chrf,meteor, rouge, ter} detect only weak lexical diversity, with the majority of delta values falling below $10\%$. In contrast, our proposed GLVS metric identifies at least $5\%$ lexical variation across all models, with most values exceeding $10\%$, indicating substantial lexical diversity. While standard metrics can detect diversity in specific cases—such as Llama-2-7b (pre-trained)~\cite{llama2} and Llama-2-70b-chat (instruct-tuned)~\cite{llama2}, they fail to do so for the majority of models. Furthermore, the trends identified by GLVS and reference-based metrics are sometimes divergent. For instance, the significant lexical diversity of Llama-2-7b-chat is captured by GLVS but remains undetected by all other metrics. This highlights the inherent insensitivity of reference-based metrics, which suffer from an over-reliance on limited gold-standard references and an inability to account for potential references caused by multimodality. Finally, GLVS values can be used to reflect the magnitude of lexical diversity across different ND-MT systems. Because deterministic MT systems score $100$ and we utilize delta values, lower GLVS scores indicate higher diversity. Besides, Although reference-based metrics are weak on detecting lexical diversity, their small delta values demonstrate that ND-MT systems generate high-quality results like D-MT. A primary advantage of GLVS is its practical utility in real-world deployment, as it eliminates the need for reference translations. In the main body, we focus on the overview analysis, and we list the concrete analysis and more results in Figure~\ref{fig:delta_lexical_std_en-zh} and  Appendix~\ref{results_greedy}.

\paragraph{ND-MT maintains the semantic equivalence}
Figure~\ref{fig:delta_semantic_mean_en-zh} illustrate the mean delta results for semantic metrics. We first find that semantic-based metrics are substantially smaller than those for lexical-based metrics (Figure~\ref{fig:delta_lexical_mean_en-zh}), with differences below 10 percentage points for most MT systems, except for specific cases like Llama-2-7b (pre-trained)~\cite{llama2} and Llama-2-70b-chat (instruct-tuned)~\cite{llama2}. Apart from that, the Std delta results from Figure~\ref{fig:delta_semantic_std_en-zh} almost remain below 10 percentage points across all metrics, demonstrating the MT systems maintain strong semantic equivalence under non-deterministic settings. In-depth discussion and baseline results are provided in Appendix~\ref{results_greedy}.

\paragraph{ND-MT has the potential to provide better candidates than D-MT} 
We further investigate the capacity of ND-MT systems to generate superior-quality candidates. Specifically, we identify the best-performing candidate within each group according to each metric and compute the average maximum scores across the dataset. It is important to note that in real-world deployment, reference translations are typically unavailable; consequently, practical systems must rely on selection strategies such as Minimum Bayes Risk (MBR) decoding~\cite{muller-sennrich-2021-understanding} with auxiliary ranking functions~\cite{gonzalez-rubio-casacuberta-2013-improving}. Therefore, these \textit{Max Delta} results should be interpreted as the theoretical potential of ND-MT to produce high-quality translations. As illustrated in Figure~\ref{fig:delta_max_en-zh}, ND-MT systems demonstrate a consistent and substantial performance gain on both lexical and semantic, revealing their capacity to generate higher-quality outputs than deterministic baselines. Furthermore, our findings validate the existence of a high quality upper bound, suggesting that deterministic selection methods (like MBR) have significant room to improve final translation results.

\paragraph{Generality of ND-MT in Addressing Multimodality}
Finally, we evaluate the generality of ND-MT potential across different language pairs. We test ⟨German, English⟩ and ⟨Russian, English⟩ in both directions with five state-of-the-art LLM-based MT models~\cite{llama2,qwen2.5,qwen3}. The results in Figure~\ref{fig:delta_all_en-de-ru} exhibit similar trends to those observed in Figure~\ref{fig:delta_all_en-zh_mean}, leading us to conclude that modern ND-MT systems demonstrate significant potential for generating diverse candidates while maintain semantic equivalence, effectively addressing multimodality limitations. Our experimental evidence indicates that modern MT systems learn translation through semantic equivalence and lexical diversity, positioning them as viable alternatives to D-MT systems. Future research can unlock the full potential of ND-MT systems in generating higher-quality translation candidates.


\subsection{Temperature Constraints on ND-MT}\label{temperature-constraint}
While we have demonstrated the potential of ND-MT in addressing multimodality challenges, the quality of generated candidates depends critically on the temperature parameter. We further investigate the effect of temperature on the performance of ND-MT. Unlike previous fine-grained studies aimed at identifying optimal parameters for generating the best single candidate, we examine how temperature influences the overall potential of ND-MT. We conduct experiments on WMT23 EN-ZH using five models~\cite{llama2,qwen2.5,qwen3}, with qwen3-chat-8 serving as a representative example, as all models exhibit similar trends. We list all the results and detailed analysis for metrics and models in Appendix~\ref{temp_effect_all}

\begin{figure}[!ht]
    \centering
    \begin{subfigure}[b]{0.45
    \linewidth}
        \centering
        \includegraphics[width=\textwidth]{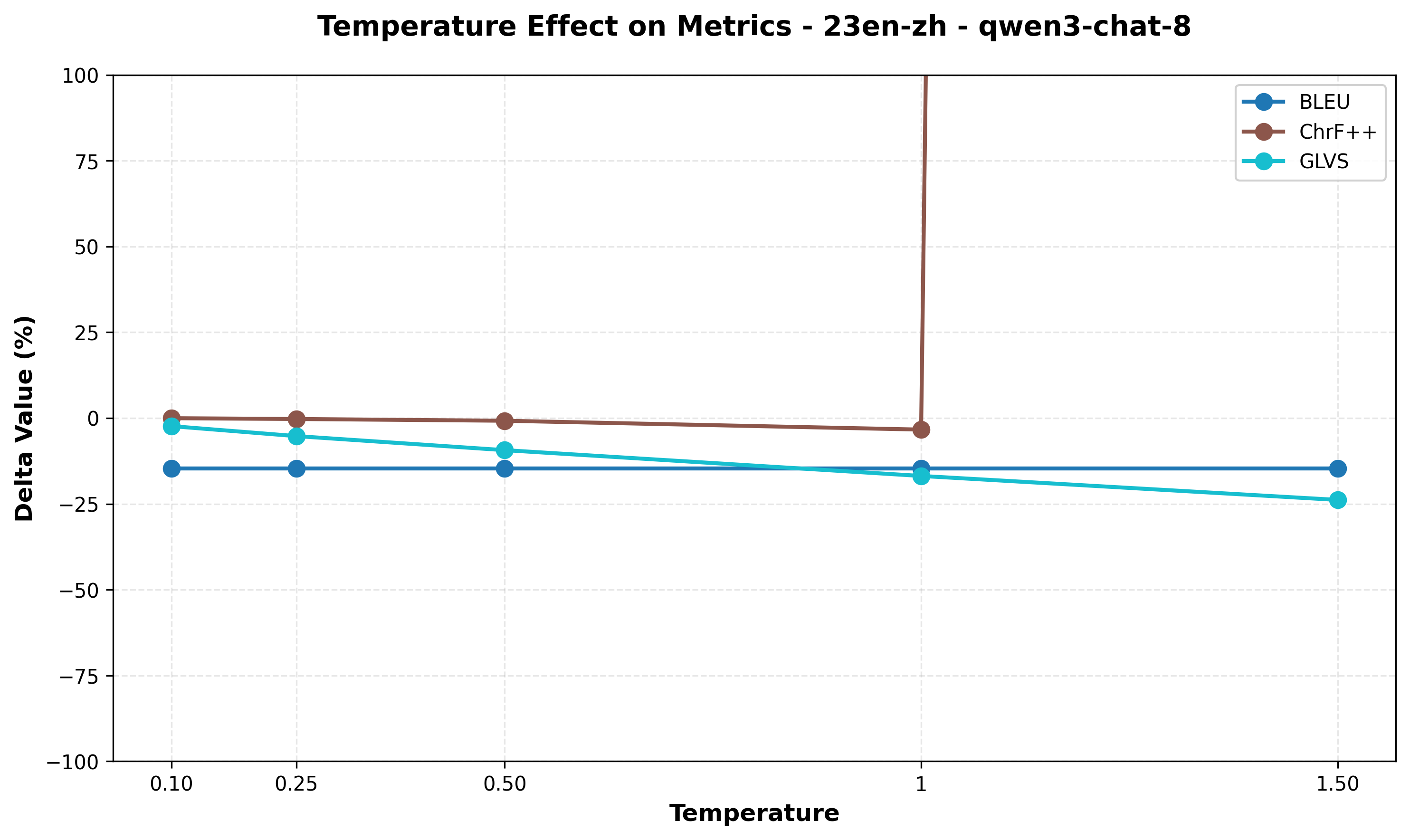}
        \caption{Temperature Effect on Lexical Metrics}
        \label{fig:temp_qwen3-chat-23en-zh-lexical}
    \end{subfigure}
    \vspace{0.1cm}
    \begin{subfigure}[b]{0.45\linewidth}
        \centering
        \includegraphics[width=\textwidth]{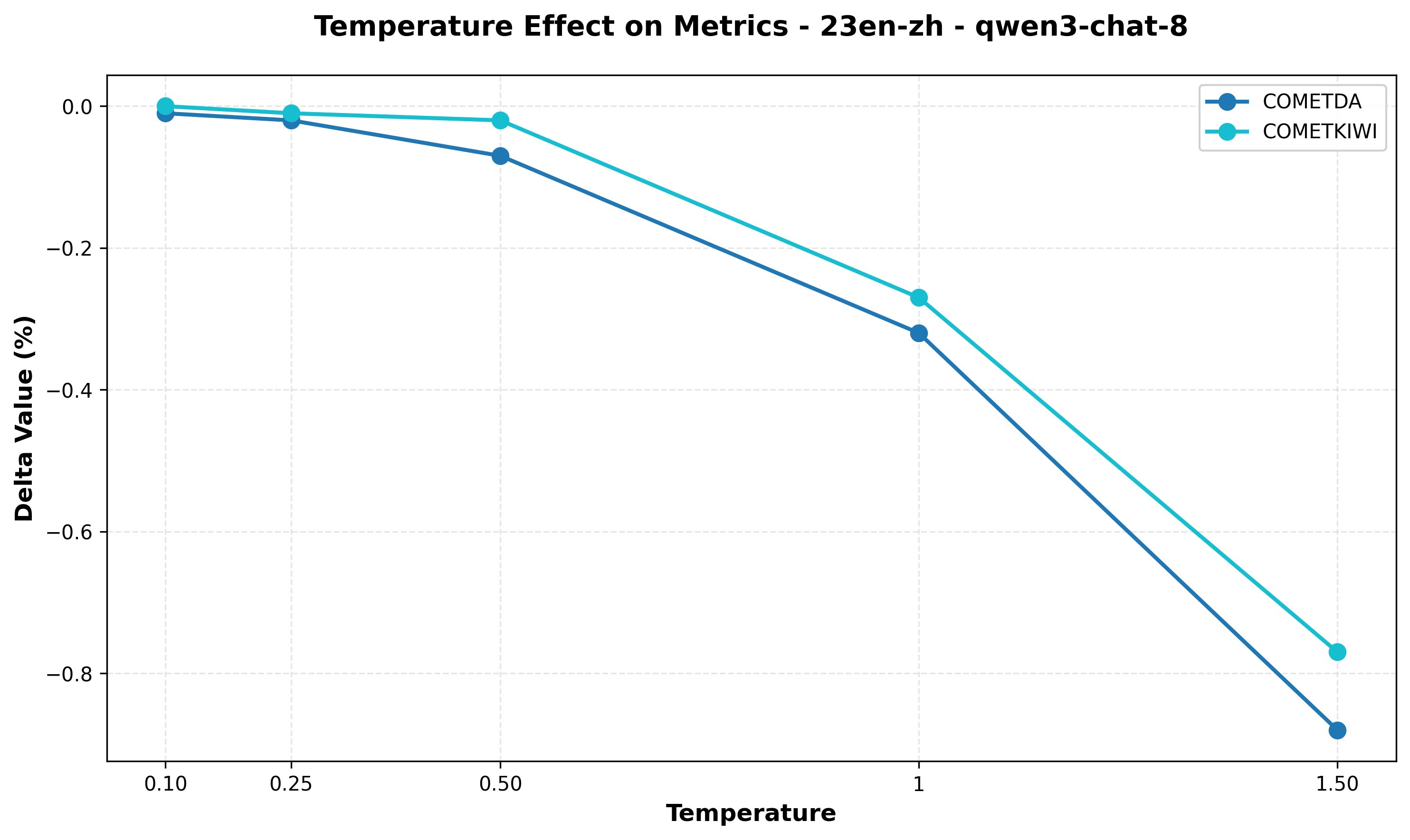}
        \caption{Temperature Effect on Semantic Metrics}
        \label{fig:temp_qwen3-chat-23en-zh-semantic}
    \end{subfigure}
\caption{The temperature effect for qwen3-chat-8 model on WMT23 EN-ZH dataset on GLVS, BLEU~\cite{bleu}, ChrF++~\cite{bleu} of lexical metrics and COMETDA~\cite{comet20da}, COMETKIWI~\cite{comet22kiwi} of semantic metrics.}
    
    \label{fig:temp_constraint}
\end{figure}

We evaluate both lexical diversity and semantic equivalence using the same metrics from Section~\ref{eval_methods}. For lexical analysis, we select GLVS as a reference-free metric, and BLEU~\cite{bleu} and ChrF++~\cite{chrf} as reference-based metrics that measure effects at the lexical and character levels, respectively. For semantic analysis, we choose COMETDA~\cite{comet20da} and COMETKIWI~\cite{comet22kiwi} as reference-based and reference-free metrics, respectively. Figure~\ref{fig:temp_constraint} presents the results. GLVS shows a decreasing trend as temperature increases, indicating that lexical diversity grows with temperature, which aligns with the general purpose of raising temperature: making a broader range of lexical items more probable. Notably, ChrF++ exceeds 100 at higher temperatures, indicating its unreliability for evaluation on ND-MT. The semantic metrics exhibit a monotonic decreasing trend, indicating that \textbf{as temperature increases, ND-MT maintains lexical diversity while sacrificing semantic equivalence}. In practical applications, the acceptable degree of semantic degradation depends on the specific use case and the baseline semantic quality. Our observations align with previous findings that non-deterministic systems show weaker performance than deterministic systems on certain downstream tasks~\cite{song-etal-2025-good}. 

In summary, to harness the potential of ND-MT, temperature values must be carefully calibrated to maintain both lexical diversity and semantic equivalence when addressing multimodality challenges. Additionally, the effects of specific temperature settings should be evaluated in advance to align with application requirements. Our experimental evidence reveals that semantic equivalence decreases while lexical diversity increases with rising temperature, providing valuable guidance for determining optimal temperature configurations in future ND-MT research and applications.

\section{The Under-Explored Space of ND-MT on the Evaluation Scheme}

\subsection{Limitations of the Current D-MT Evaluation Scheme on ND-MT}
In Section~\ref{temperature-constraint}, we demonstrate the potential of ND-MT to address multimodality challenges by providing lexically diverse candidates while maintaining semantic equivalence within the candidate set. This raises an important question: \textbf{\textit{how should we evaluate current and future ND-MT systems?}} The prevailing generate-once evaluation paradigm relies on established metrics that have been validated through human assessment. However, this paradigm is primarily suited for D-MT for two key reasons: 1) The multimodality challenge represents a fundamental limitation that affects both the design and measurement capabilities of existing metrics. For instance, lexical-based metrics such as BLEU and ChrF++ allow multiple references during evaluation, yet this assumes the availability of such references, which is often impractical to obtain. Conversely, semantic-based metrics leverage large-scale supervised training to mitigate multimodality issues. However, their effectiveness remains constrained by the scale of the training data and computational resources. 2) Evaluating ND-MT systems using humans is increasingly impractical. Because these systems generate a high volume of translation candidates, and because those outputs change significantly depending on the temperature setting, researchers would need to conduct a massive number of evaluations to reach a reliable conclusion. This creates a \textit{bottleneck} where human assessment becomes far too slow and expensive to be sustainable.

In this section, we investigate the under-explored domain of ND-MT evaluation frameworks. First, we examine an intuitive approach that directly applies evaluation rankings from D-MT, which reveals significant inconsistencies. Second, we evaluate current metrics using group-based measurements and identify the Buckets Effect in ND-MT that influences ranking determination. Finally, we propose the \textit{ExpectoSample} strategy to identify reliable metrics for selecting robust ND-MT systems.

\subsection{The Inconsistent Evaluation Results between ND-MT and D-MT}
One intuitive approach is to directly apply the ranking from deterministic MT systems to their non-deterministic counterparts. We evaluate this approach by computing Spearman's $\rho$ and Kendall's $\tau$ across five aggregation methods: \emph{min}, \emph{max}, \emph{mean}, \emph{random}, and \emph{std}. Specifically, we hypothesize that higher-ranked MT systems possess stronger capabilities for generating high-quality candidates; consequently, we expect \emph{std} to exhibit high negative correlation (i.e., higher-ranked MT systems should produce lower \emph{std} values).

The results in Tables~\ref{tab:corr_lexical} and \ref{tab:metrics_corr_sample_semantic} present correlations for lexical-based and semantic-based metrics, respectively. While most metrics demonstrate moderate to strong correlations exceeding 0.5 for both Spearman's $\rho$ and Kendall's $\tau$ (with TER being a notable exception), the observed gaps suggest that D-MT evaluation rankings provide limited reliability when applied to ND-MT systems. Furthermore, the weak correlations for \emph{std} suggest that assessing the robustness of ND-MT systems requires evaluation frameworks that extend beyond traditional deterministic approaches.
\begin{table}[t]
\centering
\caption{Correlation Results of Lexicon-based Metrics on WMT23 EN-ZH for 22 ND-MT Systems.}
\label{tab:corr_lexical}
\scriptsize
\setlength{\tabcolsep}{3pt}
\begin{tabular}{@{}lccccc@{}}
\toprule
Strategy & BLEU & METEOR & ROUGE & TER & ChrF++ \\
\midrule
\multicolumn{6}{@{}l}{\textit{Kendall's $\tau$ / p-value}} \\
\midrule
Min    & .69/.00 & .68/.00 & .70/.00 & .19/.22 & .69/.00 \\
Max    & .69/.00 & .70/.00 & .71/.00 & .27/.08 & .70/.00 \\
Mean   & .67/.00 & .68/.00 & .69/.00 & .32/.04 & .72/.00 \\
Random & .69/.00 & .69/.00 & .68/.00 & .18/.26 & .71/.00 \\
Std    & -.09/.57 & -.47/.00 & -.56/.00 & .30/.05 & -.02/.91 \\
\midrule
\multicolumn{6}{@{}l}{\textit{Spearman's $\rho$ / p-value}} \\
\midrule
Min    & .87/.00 & .87/.00 & .87/.00 & .28/.21 & .87/.00 \\
Max    & .86/.00 & .87/.00 & .88/.00 & .34/.12 & .86/.00 \\
Mean   & .86/.00 & .87/.00 & .87/.00 & .33/.13 & .88/.00 \\
Random & .87/.00 & .87/.00 & .86/.00 & .20/.37 & .88/.00 \\
Std    & -.13/.57 & -.60/.00 & -.70/.00 & .35/.11 & .00/.99 \\
\bottomrule
\end{tabular}
\end{table}

\subsection{Buckets Effect of ND-MT}
\begin{table}[!ht]
\centering
\caption{Correlation Analysis of MT Evaluation Metrics Across Sampling Sizes with Lexical Metrics on WMT23 EN-ZH with Five SOTA ND-MT Systems}
\label{tab:metrics_corr_sample_lexical}
\footnotesize
\setlength{\tabcolsep}{3pt}
\begin{tabular}{@{}clcccccc@{}}
\toprule
\multirow{2}{*}{\shortstack{Size}} & \multirow{2}{*}{\shortstack{strategy}} & \multicolumn{2}{c}{BLEU} & \multicolumn{2}{c}{GLVS} & \multicolumn{2}{c}{ChrF++} \\
\cmidrule(lr){3-4} \cmidrule(lr){5-6} \cmidrule(lr){7-8}
& & $\rho$ & $\tau$ & $\rho$ & $\tau$ & $\rho$ & $\tau$ \\
\midrule
\multirow{5}{*}{20} 
& Max    & .70 & .60 & 1.0 & 1.0 & .90 & .80 \\
& Mean   & .90 & .80 & .90 & .80 & 1.0 & 1.0 \\
& Min    & \textcolor{red}{\textbf{1.0}} & \textcolor{red}{\textbf{1.0}} & \textcolor{red}{\textbf{1.0}} & \textcolor{red}{\textbf{1.0}} & \textcolor{red}{\textbf{1.0}} & \textcolor{red}{\textbf{1.0}} \\
& Rand.  & .90 & .80 & .90 & .80 & 1.0 & 1.0 \\
& Std    & .70 & .60 & .90 & .80 & 1.0 & 1.0 \\
\midrule
\multirow{5}{*}{50}
& Max    & .70 & .60 & .90 & .80 & .90 & .80 \\
& Mean   & .90 & .80 & .90 & .80 & 1.0 & 1.0 \\
& Min    & \textcolor{red}{\textbf{1.0}} & \textcolor{red}{\textbf{1.0}} & \textcolor{red}{\textbf{1.0}} & \textcolor{red}{\textbf{1.0}} & \textcolor{red}{\textbf{1.0}} & \textcolor{red}{\textbf{1.0}} \\
& Rand.  & .90 & .80 & .90 & .80 & 1.0 & 1.0 \\
& Std    & .70 & .60 & 1.0 & 1.0 & .90 & .80 \\
\bottomrule
\end{tabular}
\begin{tablenotes}
\footnotesize
\item $\rho$ = Spearman's correlation; $\tau$ = Kendall's tau. All correlations significant at $p < 0.10$.
\end{tablenotes}
\end{table}

To further investigate reliable evaluation frameworks, we conduct experiments across different sampling sizes ($\{10, 20, 50\}$) while maintaining constant temperature values for five state-of-the-art ND-MT models. For evaluation metrics, we employ BLEU, ChrF++, and GLVS as lexical-based metrics, and COMETDA and COMETKIWI as semantic metrics. The results are presented in Tables~\ref{tab:metrics_corr_sample_lexical} and \ref{tab:corr_semantic_combined}. A key observation is the \textbf{\textit{Buckets Effect}}: the worst-cases of ND-MT systems determine the system ranking across all sampling sizes and metrics. Our findings demonstrate that a controlled sampling sizes rather than arbitrarily large samples—can yield reliable evaluations with existing metrics. However, the Buckets Effect indicates the difficult for evaluation the real system performance of ND-MT since the worst-case is unknown in advance and hard to find. However, we notice that finding reliable metrics may possible but need careful filtering (concrete discussed about the metrics in Appendix~\ref{Buckets_Effect}).

\subsection{ExpectoSample: Identifying Reliable Metrics For Selecting Robust Systems}

The Buckets Effect presents significant challenges for the reliable evaluation and practical deployment of ND-MT systems. However, our empirical analysis (Tables~\ref{tab:metrics_corr_sample_lexical} and \ref{tab:corr_semantic_combined}) reveals that certain robust metrics, specifically ChrF++~\cite{chrf}, COMET-20-DA~\cite{comet20da}, and COMET-22-Kiwi~\cite{comet22kiwi} maintain consistent system rankings under both \emph{mean} and \emph{random} sampling settings. This consistency is essential for real-world since the computation cost is limited for sampling evaluation.

Motivated by these findings, we propose \textbf{ExpectoSample}, a two-stage framework designed to first filter reliable metrics and subsequently select stable ND-MT systems. The framework is built on the principle that a truly reliable metric should produce consistent system rankings regardless of sample size, while a robust ND-MT system should maintain stable performance across varying sample counts. The ExpectoSample strategy consists of the following two steps:
\textbf{1) Metric Reliability Filtering:} We examine the system-ranking correlations across a set of increasing sampling sizes $\mathcal{X} = \{X_1, X_2, X_3\}$, where $2X_1 \le X_2$ and $2X_2 \le X_3$. Given a set of ND-MT systems and candidate metrics, we retain only those metrics whose \emph{mean} ranking correlation (e.g., Spearman's $\rho$) remains above a stability threshold $\epsilon$ (where $0 < \epsilon \le 1$) across all sampling pairs. \textbf{2) System Selection and Deployment:} Utilizing the filtered reliable metrics from Step 1, we rank the available ND-MT systems using a small sample size $X_k$ (typically $X_k \le 10$). The top-$k$ performing systems are then identified as the most reliable candidates for production-level usage. 

Notably, the set of ND-MT systems used for metric filtering in Step 1 does not need to be identical to the systems evaluated in Step 2. This decoupling enhances the generalizability of the ExpectoSample strategy, as metrics proven reliable on a reference benchmark can be confidently applied to rank new or unseen systems in real-world deployment.

\section{Discussion and Future Directions}

\paragraph{Exploring the Temperature-Constrained Nature of ND-MT}
While we focused on temperature due to its accessibility in API-based models, other parameters like \emph{top-k}~\cite{DBLP:journals/corr/abs-2505-19371} and \emph{top-p}~\cite{DBLP:conf/iclr/HoltzmanBDFC20} warrant further investigation to achieve a fine-grained understanding of non-determinism. This study establishes a general framework for assessing ND-MT potential, though future work should evaluate how diverse sampling strategies interact with these findings.

\paragraph{Upper Bounds for MBR and Re-ranking Strategies}
Our results highlight a significant performance gap between mean and maximum candidate quality, providing a theoretical justification for the success of Minimum Bayes Risk (MBR)~\cite{muller-sennrich-2021-understanding} and re-ranking methods~\cite{gonzalez-rubio-casacuberta-2013-improving}. While these selection techniques aim to capture high-quality candidates, our analysis of the Buckets Effect suggests that their success is inherently bounded by the underlying system's generation capability.

\paragraph{Limitations in Automated and Human Evaluation}
We omitted LLM-as-a-Judge~\cite{kim-2025-rubric,kocmi-federmann-2023-gemba} due to inherent biases of LLMs on preferring LLMs' generations~\cite{DBLP:conf/iclr/YeWHCZMGG0CC025}, relying instead on established lexical and semantic metrics to ensure objective evaluation. Future research can develop cost-effective human-in-the-loop~\cite{schroeder-etal-2025-just} protocols that can reliably assess non-deterministic outputs without the prohibitively high costs of full human assessment.

\paragraph{The Buckets Effect: A Metric for System-Level Improvement}
The Buckets Effect demonstrates that ND-MT performance is governed by systemic differences and vocabulary-level lower bounds rather than isolated successes on specific cases. This phenomenon suggests that meaningful progress in MT, such as through RLHF~\cite{DBLP:conf/nips/Ouyang0JAWMZASR22, DBLP:journals/ail/Lee25} or instruction tuning~\cite{rios-2025-instruction} should be measured by shifts in the entire performance distribution rather than single-output improvements.

\section{Conclusion}
In this work, we systematically investigate the potential and evaluation challenges of ND-MT. Our findings reveal their significant potential to address the long-standing multimodality challenge in MT by generating candidates with pronounced lexical diversity while maintaining semantic equivalence under specific temperature constraints. Furthermore, we identify critical evaluation challenges unique to ND-MT, specifically the inconsistency of system rankings compared to deterministic counterparts and the emergence of the \textit{Buckets Effect}. We propose \textit{ExpectoSample}, a robust strategy designed to filter for reliable evaluation metrics and identify robust ND-MT systems. Ultimately, this research provides both empirical evidence and a methodological framework for the more rigorous assessment and deployment of ND-MT.

\section*{Acknowledgments}
This work is supported by the Early Career Scheme (No.CityU 21219323) and the General Research Fund (No.CityU 11220324) of the University Grants Committee (UGC), the NSFC Young Scientists Fund (No.9240127), the Donation for Research Projects (No.9229164 and No.9229216). Additional support is provided by the Research Grants Council of the Hong Kong SAR under Grant GRF 11217823, 11216225, and Collaborative Research Fund C1042-23GF. This work is also funded by the National Natural Science Foundation of China under Grant 62371411 and the InnoHK initiative, the Government of the HKSAR, Laboratory for AI-Powered Financial Technologies.

\section*{Limitations}
While our work provides a systematic investigation into ND-MT, several limitations warrant acknowledgment. First, our experiments focus primarily on SOTA modern MT systems mainly on open-sourced models, and our findings may not generalize to other types of MT systems like closed-source MT systems. Second, our temperature analysis is constrained to a specific range of values, and the optimal temperature settings may vary across different model families, language pairs, or domain-specific applications. Third, our evaluation framework relies on existing automatic metrics (both lexical and semantic), which themselves have known limitations in capturing nuanced aspects of translation quality, such as cultural appropriateness, style consistency, and accuracy in domain-specific terminology.

Additionally, while we propose the ExpectoSample strategy for identifying reliable metrics and robust systems, our experiments are limited to sampling sizes of $\{10, 20, 50\}$. Larger sampling sizes or different sampling strategies might reveal additional patterns or insights. Furthermore, our analysis of the Buckets Effect and ranking consistency does not include human evaluation due to the impracticality of assessing numerous candidates across multiple systems and sampling sizes. Human judgment would provide valuable validation of our automatic evaluation findings, particularly regarding whether the lexical diversity we observe translates to genuinely useful translation alternatives for end users. Apart from that, our investigation covers six language directions, which, while diverse, represent only a fraction of the world's languages, and our findings may not fully capture the challenges specific to low-resource languages or linguistically distant language pairs. Finally, while the Buckets Effect is robustly validated through empirical evidence, establishing a formal theoretical framework remains a necessary step for its broader generalization.
\section*{Ethical Statement}
Our research on non-deterministic machine translation may raise several ethical considerations that warrant careful attention. First, the non-deterministic nature of ND-MT systems, which generate multiple diverse candidates for a single source sentence, introduces potential risks in high-stakes applications such as legal document translation, medical information dissemination, or official communications. While lexical diversity can be beneficial in creative or informal contexts, deploying ND-MT systems without appropriate safeguards in critical domains could lead to inconsistent or ambiguous translations that may have serious consequences. Additionally, we use open-source LLMs that may inadvertently generate outputs containing personal information from their training data. We emphasize that practitioners must carefully assess the suitability of ND-MT for their specific use cases and implement appropriate quality control mechanisms.

Second, the temperature-constrained nature of ND-MT systems presents transparency challenges. Users of MT systems may not be aware that different temperature settings can significantly affect translation quality and semantic equivalence. This lack of transparency could undermine user trust, particularly when systems produce semantically divergent outputs at higher temperatures. Developers deploying ND-MT systems have a responsibility to clearly communicate these limitations to end users and provide appropriate controls or defaults that prioritize semantic accuracy. Additionally, the evaluation challenges we identify—particularly the unreliability of traditional D-MT evaluation schemes for ND-MT—highlight the need for careful system comparison and selection. Misleading performance claims based on inappropriate evaluation methods could harm users who rely on MT systems for important communications.

Finally, we acknowledge that our released code, data, and evaluation results could potentially be misused to develop MT systems without adequate quality assurance or to make unfounded claims about system capabilities. We encourage researchers and practitioners who utilize our resources to do so responsibly, with appropriate consideration for the limitations we have identified and the potential impacts on end users across diverse linguistic and cultural communities.

\bibliography{custom}

\clearpage
\appendix

\section{Model Statistics and Implementation}\label{model_statistics}
We summarize the evaluated systems in Table~\ref{tab:models_fixed}. The table specifies the model family, parameter count, and architecture type. Additionally, we distinguish between model variants (Base, Chat, Reasoning) to indicate the corresponding machine translation prompts applied during inference. Apart from that, all models were obtained from the Hugging Face Hub under open-source licenses.\footnote{\url{https://huggingface.co/models}} Evaluation metrics were implemented using the Hugging Face \texttt{evaluate} library.\footnote{\url{https://github.com/huggingface/evaluate}} Specifically, for the semantic metrics, we utilize the \texttt{Unbabel/wmt22-comet-da}\footnote{\url{https://huggingface.co/Unbabel/wmt22-comet-da}} and \texttt{Unbabel/wmt22-cometkiwi-da}\footnote{\url{https://huggingface.co/Unbabel/wmt22-cometkiwi-da}} checkpoints to represent COMETDA~\cite{comet20da} and COMETKIWI~\cite{comet22kiwi}, respectively.
\begin{table}[t]
    \centering
    \small
    \begin{tabularx}{\linewidth}{l X r c}
        \toprule
        \textbf{Model} & \textbf{Variant} & \textbf{Params (B)} & \textbf{Type} \\
        \midrule
        \multicolumn{4}{l}{\textit{Encoder-Decoder Models}} \\
        \midrule
        mBART     & NMT & 0.68 & Dense \\
        NLLB-200  & NMT & 0.6, 3.3, 54 & Dense \\
        \midrule
        \multicolumn{4}{l}{\textit{Decoder-Only: Llama Family}} \\
        \midrule
        Llama 2   & Base & 7, 70 & Dense \\
        Llama 2   & Chat & 7, 70 & Dense \\
        Llama 3   & Base & 8, 70 & Dense \\
        Llama 3   & Chat & 8, 70 & Dense \\
        \midrule
        \multicolumn{4}{l}{\textit{Decoder-Only: Qwen Family}} \\
        \midrule
        Qwen 2.5  & Base & 7, 72 & Dense \\
        Qwen 2.5  & Chat & 7, 72 & Dense \\
        Qwen 3    & Chat & 8 & Dense \\
        Qwen 3    & Reasoning & 8 & Dense \\
        \midrule
        \multicolumn{4}{l}{\textit{Decoder-Only: DeepSeek Family}} \\
        \midrule
        DeepSeek (Llama) & Reasoning & 8 & Dense \\
        DeepSeek (Qwen)  & Reasoning & 7 & Dense \\
        DeepSeek-R1      & Reasoning & 671 & MoE \\
        \midrule
        \multicolumn{4}{l}{\textit{Other Decoder-Only Models}} \\
        \midrule
        MiniCPM   & Chat & 16 & MoE \\
        \bottomrule
    \end{tabularx}
    \par\noindent \textbf{Abbreviations:} \textbf{Base}: Pre-trained only; \textbf{SFT}: Instruction-tuned; \textbf{Reasoning}: Reinforcement Learning tuned.
    \caption{Specifications of language models used in our experiments. We report the model family, specific training variant (Base, Chat, or Reasoning), parameter counts, and the underlying architecture type (standard Dense versus Mixture-of-Experts models).}
    \label{tab:models_fixed}
\end{table}

\section{Prompting Strategies}
To ensure fair evaluation across varying architectures, we tailor our prompting strategies to the specific training stage of each model, as detailed in Table~\ref{tab:prompts}.
\begin{table*}[t]
    \centering
    \small
    \begin{tabularx}{\linewidth}{l l X}
        \toprule
        \textbf{Variant} & \textbf{Strategy} & \textbf{Input Template / Format} \\
        \midrule
        \textbf{NMT} & Language Tokens & \texttt{<src\_lang\_token> [Source Sentence]} \\
         & & \textit{Example:} \texttt{zho\_Hans Hello world} \\
        \midrule
        \textbf{Base} & 5-Shot In-Context & 
        \begin{CJK*}{UTF8}{gbsn}
            \texttt{Translate the following English sentences to Chinese:} \newline
            \texttt{English: The weather is beautiful today.} \newline
            \texttt{Chinese: 今天天气很好。} \newline
            \texttt{... (3 other examples) ...} \newline
            \texttt{English: <text>} \newline
            \texttt{Chinese:} 
        \end{CJK*} 
        \\
        \midrule
        \textbf{Chat / Reasoning} & Zero-Shot Instruction & \texttt{Translate the following <src\_lang> text to <tgt\_lang>. Only provide the translation, no explanations:} \newline
                                                \newline
                                                \texttt{<text>} \\
        \bottomrule
    \end{tabularx}
    \caption{Overview of prompting strategies. We use standard tokens for \textbf{NMT}, a fixed 5-shot template with human-curated LLM demonstrations for \textbf{Base} models, and constrained zero-shot instructions for \textbf{Chat/Reasoning} models. In practice, generic language placeholders in the templates are instantiated with the specific translation direction (e.g., ``English'' to ``Chinese'').}
    \label{tab:prompts}
\end{table*}

\paragraph{Encoder-Decoder (NMT).}
For standard NMT systems like NLLB-200 and mBART, we adhere to the official default configurations. Specifically, we append the designated language identification tokens (e.g., \texttt{eng\_Latn}, \texttt{zho\_Hans}) to the input sequence to specify the translation direction. This tokenization process is automated using the standard preprocessing pipelines provided by the Hugging Face library~\cite{huggingface}~\footnote{\url{https://huggingface.co/}}.

\paragraph{Base Models (Pre-trained).}
For pre-trained decoder-only models, we employ \textbf{5-shot in-context learning} (ICL) to stabilize generation~\cite{wang-etal-2024-mitigating-language}. As detailed in Table~\ref{tab:prompts}, the prompt consists of five fixed parallel demonstrations (e.g., English: ``The weather is beautiful today'' $\to$ Chinese: ``\begin{CJK*}{UTF8}{gbsn}今天天气很好。\end{CJK*}''), followed by the input query. This format effectively primes the model to adhere to the expected output structure: \texttt{English: [Input] Chinese:}. To ensure high quality, the demonstration examples were manually curated from LLM-generated candidates.

\paragraph{Chat and Reasoning Models.}
For instruction-tuned (Chat) and reasoning-optimized variants (including DeepSeek-R1), we use a structured zero-shot prompt. To avoid the common issue of chat models generating conversational filler (e.g., "Sure, here is the translation..."), we explicitly constrain the output with the instruction: \textit{"Only provide the translation, no explanations."} The input is formatted as a single user message containing the source text and the target language specification.

\section{Evaluation Results on D-MT}\label{results_greedy}
\subsection{Original Evaluation Results}
We present the evaluation results for Deterministic Machine Translation (D-MT) on the WMT23 English$\to$Chinese (EN-ZH) dataset. Table~\ref{tab:dmt_original_metrics_23enzh} summarizes the performance using standard lexical-based metrics, while Table~\ref{tab:dmt_semantic_metrics_enzh} details the corresponding results for semantic-based metrics.

\paragraph{LLMs outperform traditional NMT baselines.}
As shown in Table~\ref{tab:dmt_original_metrics_23enzh}, large language models (LLMs) significantly surpass dedicated NMT systems. Among NMT baselines, mBART-50 is the strongest performer (31.41 BLEU), yet it is easily overtaken by modern 7B-scale LLMs. For instance, Llama3-8B achieves 37.60 BLEU, and Qwen2.5-7B reaches 44.00 BLEU, demonstrating that general-purpose pre-training is highly effective for translation even without task-specific architectural bias.

\paragraph{Scaling and Architecture Dominance.}
Performance scales consistently with model size. The Qwen2.5-72B model achieves state-of-the-art results across nearly all metrics, setting the benchmark at \textbf{48.49 BLEU} and \textbf{86.94 COMET}. Notably, the Qwen family consistently outperforms Llama models of comparable size (e.g., Qwen2.5-7B surpasses Llama3-8B by +6.4 BLEU), likely due to its stronger multilingual pre-training corpus.

\paragraph{The "Chat" Alignment Tax.}
Comparing Base models to their Chat variants reveals a mixed impact of instruction tuning. For the Llama 2 family, the Chat versions suffer a catastrophic performance drop (e.g., Llama2-7B drops from 28.32 to 15.39 BLEU), accompanied by exploding TER scores (756.12), indicating severe repetition or formatting issues. However, newer models like Llama 3 and Qwen 2.5 show minimal degradation—or even slight improvements—in their Chat variants, suggesting that modern alignment techniques (RLHF) have become more robust for translation tasks.

\paragraph{Reasoning Models Struggle with Form.}
Surprisingly, reasoning-optimized models (e.g., DeepSeek-R1, Qwen3-Reasoning) underperform compared to standard dense models. Despite its massive scale, DeepSeek-R1 (671B) achieves only 26.77 BLEU, lower than the 7B Base models. The semantic metrics in Table~\ref{tab:dmt_semantic_metrics_enzh} confirm this trend (COMET 81.05 vs. 86.94 for Qwen2.5-72B). This suggests that "reasoning" reinforcement learning, while powerful for logic, may introduce verbosity or structural deviations that are penalized in standard translation evaluation.
\paragraph{API Instability Impacts Reasoning Models.}
Contrary to expectations based on parameter scale, reasoning-optimized models (e.g., DeepSeek-R1, DS-Qwen) significantly underperform standard dense models. Our error analysis reveals that this is largely due to **API instability** rather than inherent model capability. We observed frequent occurrences of empty responses caused by network timeouts or API overloading during inference. These null outputs are penalized heavily by lexical metrics—resulting in low BLEU scores (e.g., 26.77 for DeepSeek-R1)—and distort semantic evaluations, highlighting the reliability challenges of deploying API-based models for large-scale benchmarks.
\begin{table}[!ht]
\centering
\caption{The Original Lexical-based Metrics Results of D-MT Models on 23 EN-ZH}
\label{tab:dmt_original_metrics_23enzh}
\scriptsize
\setlength{\tabcolsep}{2pt}
\begin{tabular}{@{}lcccccccc@{}}
\toprule
\textbf{Model} & \textbf{BLEU} & \textbf{MET} & \textbf{R-1} & \textbf{R-2} & \textbf{R-L} & \textbf{chrF} & \textbf{TER} \\
\midrule
\multicolumn{8}{c}{\textit{NMT}} \\
mBART-50    & 31.41 & 46.90 & 55.98 & 27.72 & 52.98 & 25.34 & 145.50 \\
NLLB-600M   & 26.02 & 36.81 & 48.02 & 24.31 & 45.18 & 21.03 & 108.01 \\
NLLB-3.3B   & 26.34 & 36.84 & 48.20 & 25.72 & 45.50 & 21.78 & 123.47 \\
NLLB-54B    & 24.17 & 33.72 & 45.35 & 24.43 & 42.90 & 20.44 & 111.43 \\
\midrule
\multicolumn{8}{c}{\textit{Pre-trained (7-8B)}} \\
Llama2-7B   & 28.32 & 45.71 & 56.11 & 27.11 & 52.70 & 24.54 & 102.84 \\
Llama3-8B   & 37.60 & 54.77 & 63.07 & 35.48 & 59.67 & 31.63 & 103.47 \\
Qwen2.5-7B  & 44.00 & 61.46 & 67.89 & 42.31 & 64.33 & 36.69 & \textbf{98.25}  \\
\midrule
\multicolumn{8}{c}{\textit{Pre-trained (70-72B)}} \\
Llama2-70B  & 40.53 & 58.64 & 65.89 & 39.13 & 62.29 & 33.74 & 101.76 \\
Llama3-70B  & 44.19 & 61.89 & 68.08 & 42.43 & 64.47 & 37.38 & 99.27  \\
Qwen2.5-72B & \textbf{48.49} & \textbf{65.85} & \textbf{70.80} & \textbf{46.98} & \textbf{67.62} & \textbf{40.36} & 98.38  \\
\midrule
\multicolumn{8}{c}{\textit{Chat (7-8B)}} \\
Llama2-C-7B  & 15.39 & 29.23 & 34.64 & 13.51 & 32.14 & 13.56 & 756.12$^*$ \\
Llama3-C-8B  & 35.58 & 53.24 & 61.11 & 33.77 & 57.51 & 30.00 & 108.84 \\
Qwen2.5-C-7B & 39.43 & 58.02 & 64.53 & 37.28 & 61.03 & 32.90 & 104.77 \\
Qwen3-C-8B   & 41.97 & 60.52 & 66.00 & 40.16 & 62.76 & 34.91 & 99.35  \\
\midrule
\multicolumn{8}{c}{\textit{Chat (70-72B)}} \\
Llama2-C-70B & 16.04 & 36.90 & 33.75 & 15.36 & 31.13 & 16.13 & 2169.34$^*$ \\
Llama3-C-70B & 42.13 & 59.99 & 66.07 & 40.17 & 62.78 & 35.48 & 99.09  \\
Qwen2.5-C-72B & 45.88 & 63.89 & 69.13 & 44.30 & 65.77 & 38.20 & 103.38 \\
\midrule
\multicolumn{8}{c}{\textit{Reasoning}} \\
MiniCPM-8x2  & 42.30 & 59.68 & 66.36 & 40.51 & 63.02 & 34.47 & 107.09 \\
Qwen3-R-8B   & 40.39 & 57.86 & 63.56 & 38.57 & 60.60 & 33.66 & 1551.24$^*$ \\
DS-Llama-8B  & 33.61 & 50.95 & 59.34 & 31.33 & 55.51 & 27.84 & 183.59 \\
DS-Qwen-7B   & 30.35 & 49.12 & 57.25 & 28.19 & 53.27 & 25.76 & 132.72 \\
DS-671B      & 26.77 & 43.46 & 50.87 & 24.19 & 47.99 & 23.77 & 114.91 \\
\bottomrule
\end{tabular}
\begin{tablenotes}
\scriptsize
\item MET=METEOR; R-1/2/L=ROUGE-1/2/L; chrF=ChrF++; C=Chat; R=Reason; DS=DeepSeek. 
\item $^*$Exceptionally high TER values indicate potential issues. Lower TER is better; higher is better for other metrics.
\end{tablenotes}
\end{table}

\begin{table}[t]
    \centering
    \caption{Semantic-based metrics for D-MT models (En-Zh). We report COMETKIWI (KIWI), BLEURT (BLE), BERTScore (BERT), COMETDA (CMT), LASER (LSR), LaBSE (LBS), SentTrans (SNT), and XNLI.}
    \label{tab:dmt_semantic_metrics_enzh}
    
    \setlength{\tabcolsep}{1.2pt} 
    
    \resizebox{\linewidth}{!}{
        \begin{tabular}{@{}lcccccccc@{}}
            \toprule
            \textbf{Model} & \textbf{KIWI} & \textbf{BLE} & \textbf{BERT} & \textbf{CMT} & \textbf{LSR} & \textbf{LBS} & \textbf{SNT} & \textbf{XNLI} \\
            \midrule
            \multicolumn{9}{c}{\textit{NMT}} \\
            mBART-50      & 75.24 & 58.97 & 86.32 & 80.81 & 82.44 & 84.09 & 13.00 & 97.80 \\
            NLLB-600M     & 67.08 & 52.74 & 83.58 & 75.36 & 78.41 & 77.28 & 10.75 & 97.08 \\
            NLLB-3.3B     & 66.06 & 52.11 & 83.13 & 75.72 & 75.82 & 73.78 & 10.62 & 96.07 \\
            NLLB-54B      & 63.48 & 49.76 & 81.85 & 74.75 & 72.16 & 69.06 & 9.79  & 92.69 \\
            \midrule
            \multicolumn{9}{c}{\textit{Pre-trained (7-8B)}} \\
            Llama2-7B     & 71.96 & 54.39 & 84.63 & 78.77 & 79.31 & 79.89 & 14.02 & 93.96 \\
            Llama3-8B     & 77.12 & 61.15 & 87.46 & 83.73 & 81.88 & 84.24 & 14.81 & 97.32 \\
            Qwen2.5-7B    & 79.47 & 64.03 & 88.73 & 85.97 & 82.27 & 85.18 & 15.16 & 98.03 \\
            \midrule
            \multicolumn{9}{c}{\textit{Pre-trained (70-72B)}} \\
            Llama2-70B    & 76.61 & 60.99 & 87.87 & 83.18 & 82.57 & 85.36 & 14.27 & 97.76 \\
            Llama3-70B    & 78.63 & 63.72 & 88.74 & 85.40 & 82.94 & 85.72 & 15.08 & 97.89 \\
            Qwen2.5-72B   & \textbf{80.40} & \textbf{66.25} & \textbf{89.80} & \textbf{86.94} & 82.82 & 86.15 & 14.85 & \textbf{98.32} \\
            \midrule
            \multicolumn{9}{c}{\textit{Chat (7-8B)}} \\
            L2-C-7B       & 58.58 & 35.67 & 76.58 & 64.75 & 78.35 & 77.17 & 32.93$^*$ & 76.42 \\
            L3-C-8B       & 78.08 & 59.95 & 86.68 & 84.01 & 80.69 & 83.29 & 15.34 & 97.93 \\
            Q2.5-C-7B     & 79.27 & 62.08 & 87.75 & 85.40 & 81.92 & 85.11 & 15.20 & 98.05 \\
            Q3-C-8B       & 80.70 & 63.53 & 88.35 & 86.30 & 82.31 & 85.77 & 14.23 & 98.18 \\
            \midrule
            \multicolumn{9}{c}{\textit{Chat (70-72B)}} \\
            L2-C-70B      & 57.36 & 32.23 & 70.85 & 53.04 & 71.95 & 76.19 & 41.27$^*$ & 69.03 \\
            L3-C-70B      & 80.02 & 63.42 & 88.36 & 86.07 & 81.88 & 84.79 & 14.60 & 98.03 \\
            Q2.5-C-72B    & 80.57 & 65.13 & 89.22 & 86.78 & \textbf{82.76} & \textbf{85.99} & 14.60 & 98.11 \\
            \midrule
            \multicolumn{9}{c}{\textit{Reasoning}} \\
            MiniCPM       & 78.71 & 62.86 & 88.29 & 84.94 & 82.07 & 84.74 & 13.37 & 97.92 \\
            Q3-R-8B       & 79.21 & 62.38 & 87.25 & 84.62 & 81.37 & 84.26 & 15.43 & 96.77 \\
            DS-Llama      & 75.79 & 57.68 & 85.53 & 81.64 & 80.77 & 82.21 & 13.42 & 96.21 \\
            DS-Qwen       & 74.05 & 55.00 & 84.91 & 80.43 & 81.36 & 83.10 & 16.74 & 97.55 \\
            DS-671B       & 76.54 & 54.51 & 79.85 & 81.05 & 75.02 & 77.37 & 13.66 & 92.34 \\
            \bottomrule
        \end{tabular}
    } 
    \begin{tablenotes}
        \scriptsize
        \item \textbf{Metrics:} KIWI=COMETKIWI; BLE=BLEURT; BERT=BERTScore; CMT=COMETDA; LSR=LASER; LBS=LaBSE; SNT=SentTrans.
        \item \textbf{Models:} L=Llama; Q=Qwen; DS=DeepSeek; C=Chat; R=Reasoning.
        \item $^*$Anomalous SentTrans values. Best results in \textbf{bold}.
    \end{tablenotes}
\end{table}

\subsection{Ranking Inconsistency and Metric Reliability}
\label{sec:ranking_analysis}

To assess the reliability of automated evaluation, we computed the relative rankings of all 18 models across the metrics, as detailed in \textbf{Table~\ref{tab:model_rankings}}. While top-tier models show some stability, the overall analysis reveals critical weaknesses in relying on single-generation outputs.

\paragraph{Dominance vs. Disagreement.}
At the top of the leaderboard, metrics largely align: \textbf{Qwen2.5-72B} achieves the Rank \#1 position across nearly all categories (see \textbf{Table~\ref{tab:model_rankings}a} for BLEU and \textbf{Table~\ref{tab:model_rankings}b} for XNLI), confirming its status as the current state-of-the-art. However, outside the top rank, significant contradictions emerge. For instance, the \textbf{mBART-50} baseline ranks high on semantic embedding metrics (Rank \#4 in LASER, \textbf{Table~\ref{tab:model_rankings}b}) but falls to the bottom tier on lexical overlap (Rank \#16 in BLEU, \textbf{Table~\ref{tab:model_rankings}a}). This implies that while the model captures semantic intent, its surface realization diverges from the reference, a nuance that lexical metrics punish disproportionately.

\paragraph{The Fragility of Single-Metric Evaluation.}
Crucially, no two metrics produce an identical ranking order. We observe extreme divergence in the Chat-tuned models:
\begin{itemize}
    \item \textbf{Llama2-Chat-70B} is ranked as the \textbf{best} model (Rank \#1) by SentTrans, yet is rated as the \textbf{worst} (Rank \#23) by COMET and BLEU (see \textbf{Table~\ref{tab:model_rankings}}).
    \item \textbf{NLLB-600M} is ranked \#11 in TER (better than Llama2-Chat), yet \#19 in COMET (worse than Llama2-Chat).
\end{itemize}
This misalignment underscores the danger of evaluating D-MT systems based on a single deterministic generation. Since greedy decoding represents only one point on the probability curve, it is susceptible to "lucky" or "unlucky" stylistic choices that metrics weight differently. This finding motivates our shift toward Non-Deterministic (ND-MT) evaluation to capture the model's full capability rather than a single, potentially biased output.

\begin{table*}[t]
    \centering
    \caption{Rankings of D-MT Models across Lexical and Semantic Metrics. A rank of \textbf{1} indicates the best performance (Highest Score for all metrics, except Lowest Score for TER).}
    \label{tab:model_rankings}
    
    \begin{subtable}{\linewidth}
        \centering
        \caption{Lexical Metric Rankings (Lower Rank \# is Better)}
        \small
        \setlength{\tabcolsep}{3.5pt}
        \begin{tabular}{l ccccccc}
            \toprule
            \textbf{Model} & \textbf{BLEU} & \textbf{MET} & \textbf{R-1} & \textbf{R-2} & \textbf{R-L} & \textbf{chrF} & \textbf{TER} \\
            \midrule
            \textit{NMT Baselines} & & & & & & & \\
            mBART-50      & 16 & 15 & 16 & 13 & 14 & 14 & 17 \\
            NLLB-600M     & 20 & 19 & 19 & 19 & 18 & 18 & 11 \\
            NLLB-3.3B     & 19 & 18 & 18 & 16 & 17 & 17 & 15 \\
            NLLB-54B      & 21 & 20 & 20 & 18 & 19 & 19 & 13 \\
            \midrule
            \textit{Pre-trained LLMs} & & & & & & & \\
            Llama2-7B     & 17 & 16 & 15 & 15 & 16 & 15 & 8 \\
            Llama3-8B     & 13 & 11 & 12 & 10 & 10 & 10 & 10 \\
            Qwen2.5-7B    & 5 & 5 & 4 & 4 & 4 & 5 & \textbf{1} \\
            Llama2-70B    & 9 & 7 & 7 & 7 & 7 & 8 & 7 \\
            Llama3-70B    & 4 & 4 & 3 & 3 & 3 & 4 & 4 \\
            Qwen2.5-72B   & \textbf{1} & \textbf{1} & \textbf{1} & \textbf{1} & \textbf{1} & \textbf{1} & 3 \\
            \midrule
            \textit{Chat Models} & & & & & & & \\
            L2-C-7B       & 23 & 23 & 22 & 23 & 22 & 23 & 20 \\
            L3-C-8B       & 14 & 12 & 13 & 11 & 11 & 11 & 12 \\
            Q2.5-C-7B     & 11 & 8 & 8 & 9 & 8 & 9 & 9 \\
            Q3-C-8B       & 8 & 6 & 6 & 6 & 6 & 6 & 5 \\
            L2-C-70B      & 22 & 17 & 23 & 22 & 23 & 22 & 21 \\
            L3-C-70B      & 7 & 9 & 5 & 5 & 5 & 3 & 2 \\
            Q2.5-C-72B    & 3 & 3 & 2 & 2 & 2 & 2 & 6 \\
            \midrule
            \textit{Reasoning} & & & & & & & \\
            MiniCPM       & 6 & 9 & 10 & 8 & 9 & 7 & 14 \\
            Q3-R-8B       & 10 & 10 & 11 & 12 & 12 & 13 & 19 \\
            DS-Llama-8B   & 15 & 13 & 14 & 14 & 13 & 12 & 18 \\
            DS-Qwen-7B    & 18 & 14 & 17 & 17 & 15 & 16 & 16 \\
            DS-671B       & 20 & 15 & 21 & 21 & 20 & 20 & 22 \\
            \bottomrule
        \end{tabular}
    \end{subtable}

    \vspace{0.5cm} 

    \begin{subtable}{\linewidth}
        \centering
        \caption{Semantic Metric Rankings (Lower Rank \# is Better)}
        \small
        \setlength{\tabcolsep}{2.5pt}
        \begin{tabular}{l cccccccc}
            \toprule
            \textbf{Model} & \textbf{KIWI} & \textbf{BLE} & \textbf{BERT} & \textbf{CMT} & \textbf{LSR} & \textbf{LBS} & \textbf{SNT} & \textbf{XNLI} \\
            \midrule
            \textit{NMT Baselines} & & & & & & & & \\
            mBART-50      & 15 & 13 & 14 & 15 & 4 & 12 & 18 & 8 \\
            NLLB-600M     & 19 & 19 & 19 & 19 & 18 & 19 & 19 & 11 \\
            NLLB-3.3B     & 20 & 20 & 20 & 18 & 20 & 22 & 20 & 15 \\
            NLLB-54B      & 21 & 21 & 21 & 20 & 21 & 23 & 21 & 18 \\
            \midrule
            \textit{Pre-trained LLMs} & & & & & & & & \\
            Llama2-7B     & 17 & 18 & 18 & 17 & 17 & 17 & 12 & 17 \\
            Llama3-8B     & 14 & 11 & 12 & 12 & 9 & 11 & 8 & 10 \\
            Qwen2.5-7B    & 7 & 6 & 5 & 5 & 6 & 6 & 6 & 5 \\
            Llama2-70B    & 13 & 12 & 10 & 11 & 5 & 7 & 11 & 9 \\
            Llama3-70B    & 10 & 7 & 4 & 6 & 2 & 5 & 7 & 7 \\
            Qwen2.5-72B   & \textbf{1} & \textbf{1} & \textbf{1} & \textbf{1} & 3 & 2 & 7 & \textbf{1} \\
            \midrule
            \textit{Chat Models} & & & & & & & & \\
            L2-C-7B       & 22 & 22 & 22 & 22 & 18 & 20 & 2 & 20 \\
            L3-C-8B       & 12 & 14 & 13 & 10 & 16 & 13 & 5 & 6 \\
            Q2.5-C-7B     & 8 & 10 & 8 & 7 & 8 & 8 & 6 & 4 \\
            Q3-C-8B       & 2 & 8 & 6 & 4 & 5 & 4 & 10 & 3 \\
            L2-C-70B      & 23 & 23 & 23 & 23 & 22 & 21 & \textbf{1} & 21 \\
            L3-C-70B      & 6 & 9 & 5 & 5 & 10 & 9 & 9 & 5 \\
            Q2.5-C-72B    & 3 & 3 & 2 & 2 & \textbf{1} & \textbf{1} & 9 & 4 \\
            \midrule
            \textit{Reasoning} & & & & & & & & \\
            MiniCPM       & 9 & 8 & 7 & 8 & 7 & 9 & 16 & 7 \\
            Q3-R-8B       & 8 & 9 & 11 & 9 & 11 & 10 & 3 & 13 \\
            DS-Llama-8B   & 13 & 15 & 15 & 14 & 14 & 15 & 15 & 14 \\
            DS-Qwen-7B    & 16 & 16 & 16 & 16 & 12 & 14 & 4 & 9 \\
            DS-671B       & 11 & 17 & 21 & 15 & 20 & 18 & 14 & 19 \\
            \bottomrule
        \end{tabular}
    \end{subtable}
\end{table*}

\section{Temperature Effect on ND-MT}
\label{temp_effect_all}

We analyze the impact of temperature sampling on Non-Deterministic Machine Translation (ND-MT) across five state-of-the-art systems, including the Llama 2 family (Pre/Chat-7B)~\cite{llama2} and the Qwen family (2.5-Pre/Chat-7B, 3-Chat-8B)~\cite{qwen2.5,qwen3}. 

\begin{figure}[t!]
    \centering
    \begin{subfigure}[t]{0.48\linewidth} 
        \centering
        \includegraphics[width=\linewidth]{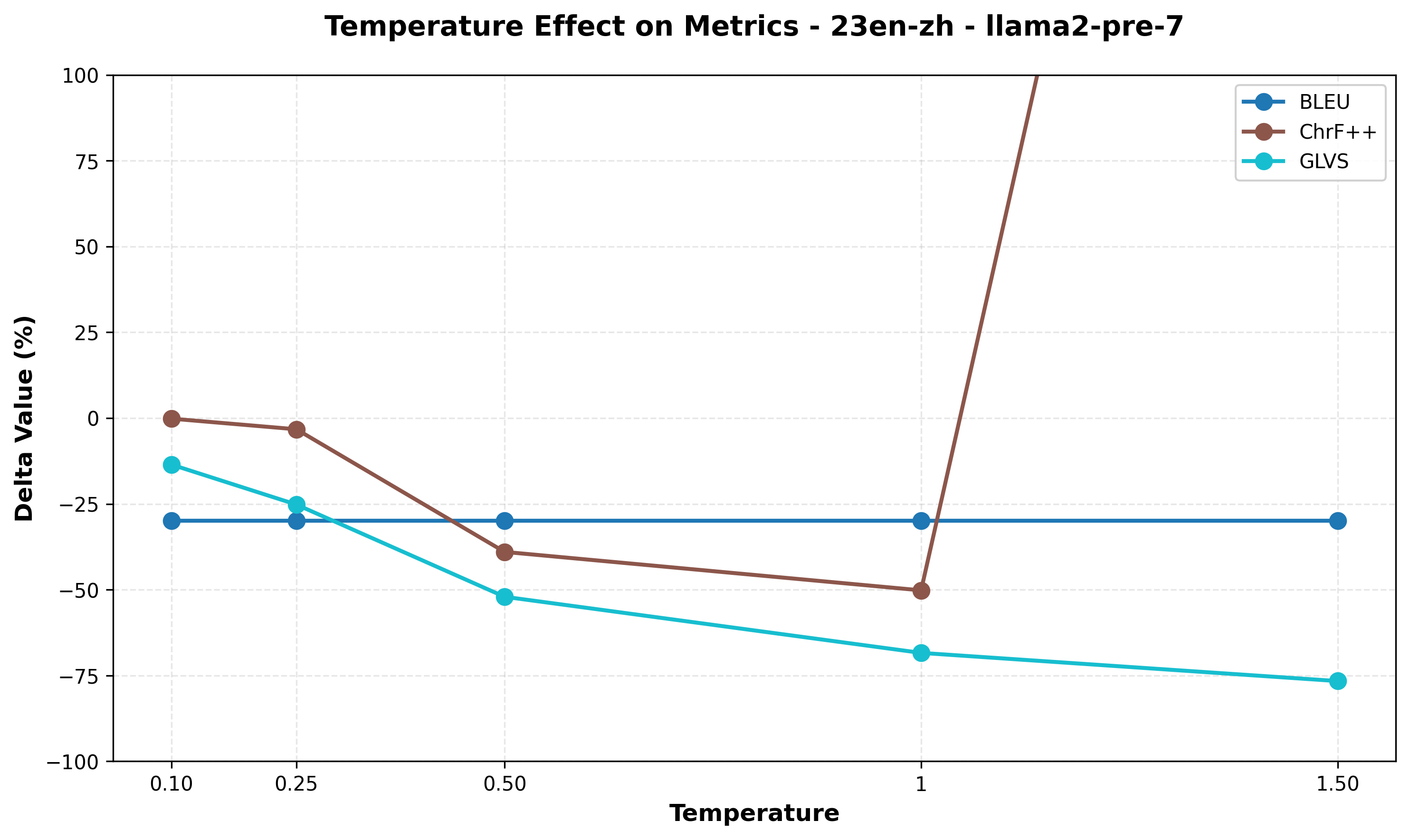}
        \caption{Llama2-Pre-7B (Lexical)}
    \end{subfigure}
    \hfill
    \begin{subfigure}[t]{0.48\linewidth} 
        \centering
        \includegraphics[width=\linewidth]{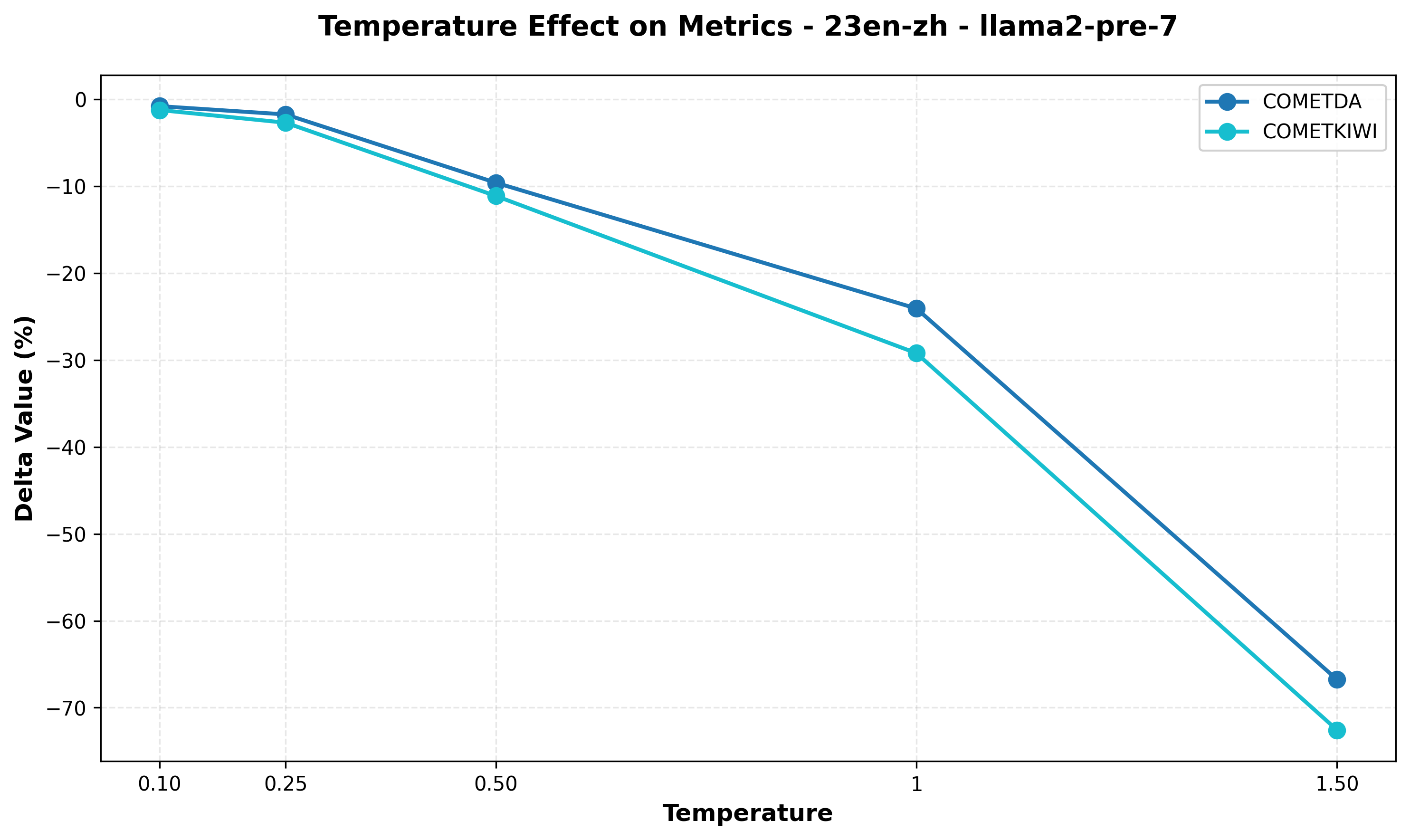}
        \caption{Llama2-Pre-7B (Semantic)}
    \end{subfigure}
    
    \vspace{0.2cm}
    
    \begin{subfigure}[t]{0.48\linewidth} 
        \centering
        \includegraphics[width=\linewidth]{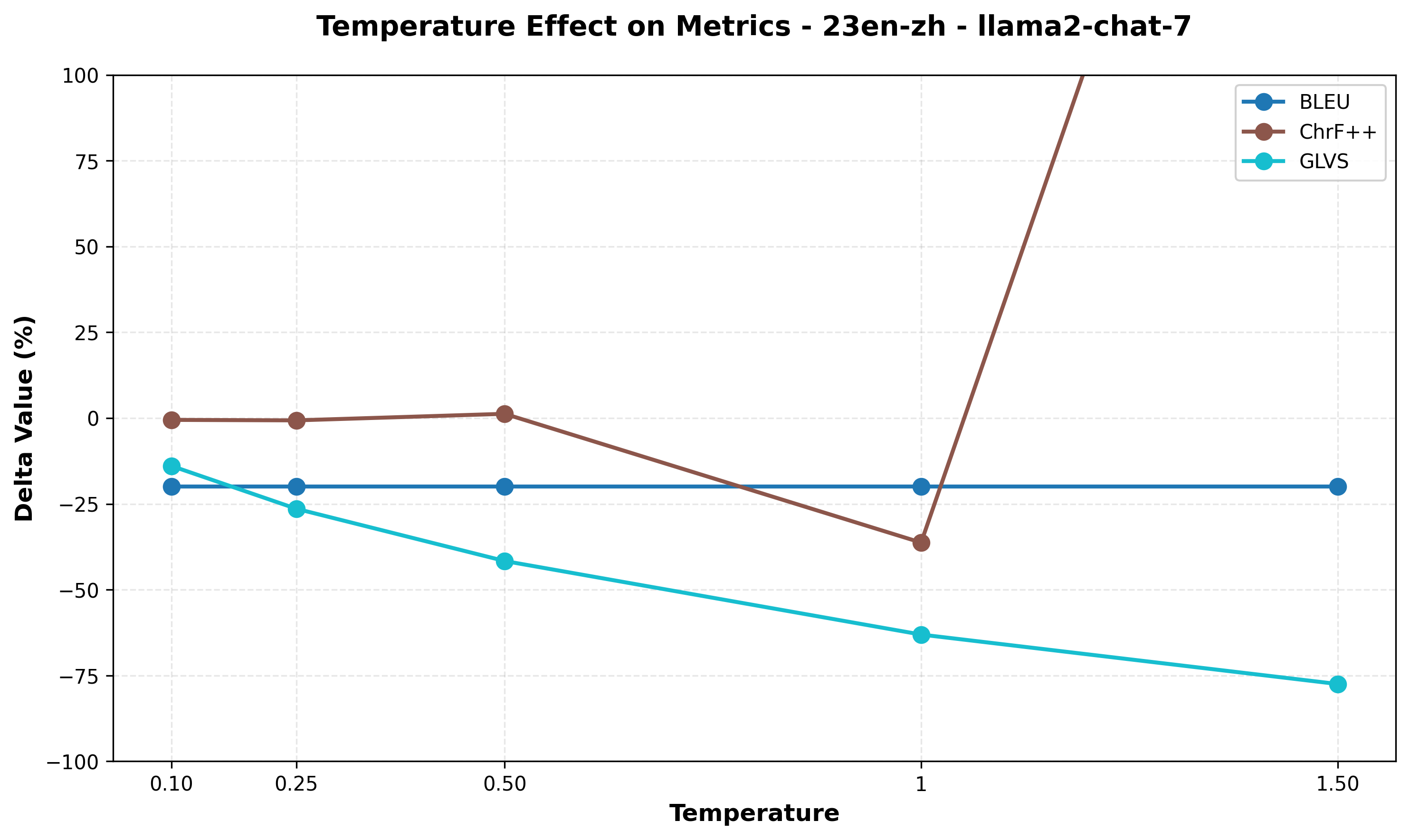}
        \caption{Llama2-Chat-7B (Lexical)}
    \end{subfigure}
    \hfill
    \begin{subfigure}[t]{0.48\linewidth} 
        \centering
        \includegraphics[width=\linewidth]{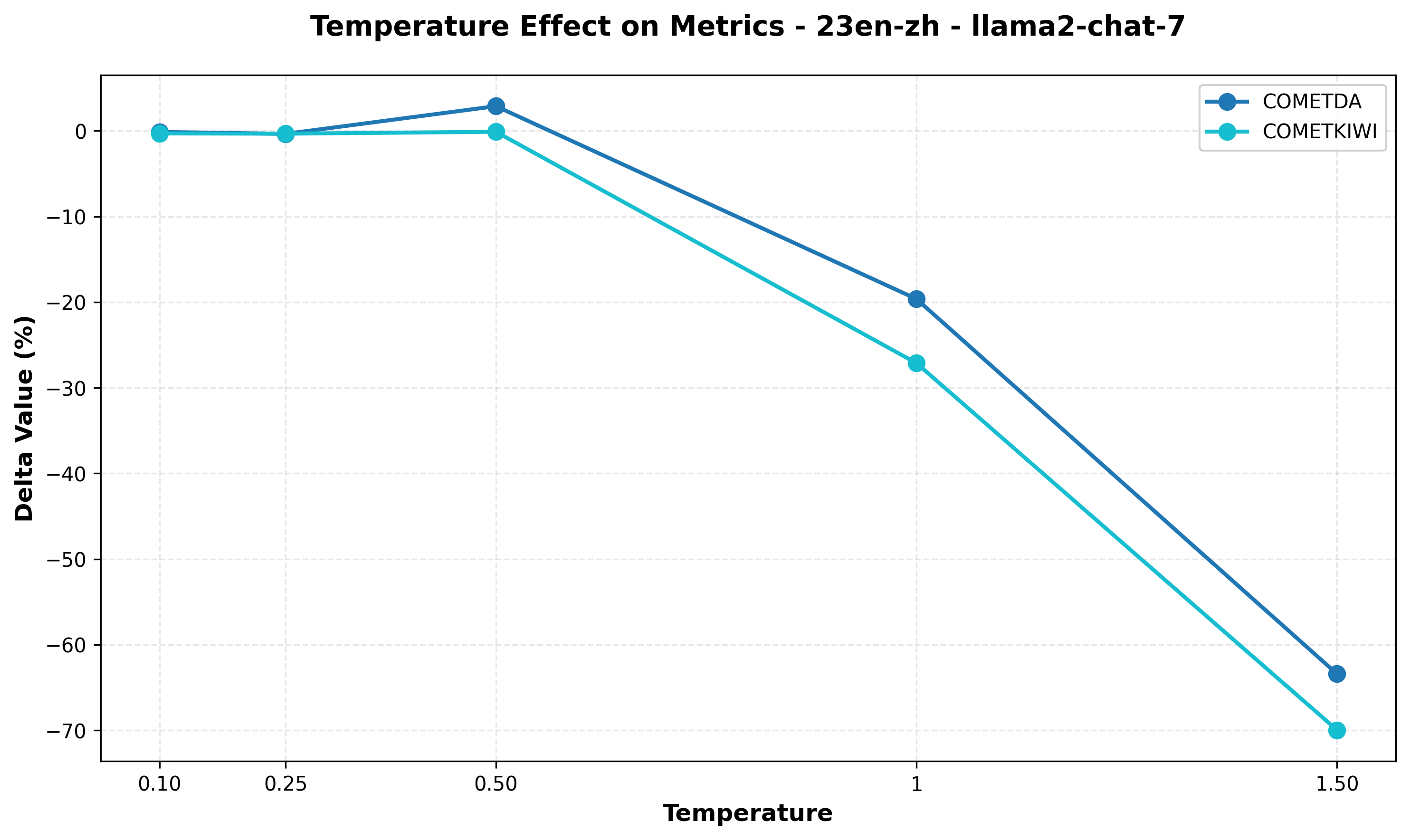}
        \caption{Llama2-Chat-7B (Semantic)}
    \end{subfigure}

    \caption{Temperature effect on Llama 2 models. While lexical metrics (left) remain stable, semantic metrics (right) show degradation at high temperatures.}
    \label{fig:temp_llama_family}
\end{figure}

\begin{figure*}[t!] 
    \centering
    \begin{subfigure}[b]{0.48\linewidth}
        \centering
        \includegraphics[width=\linewidth]{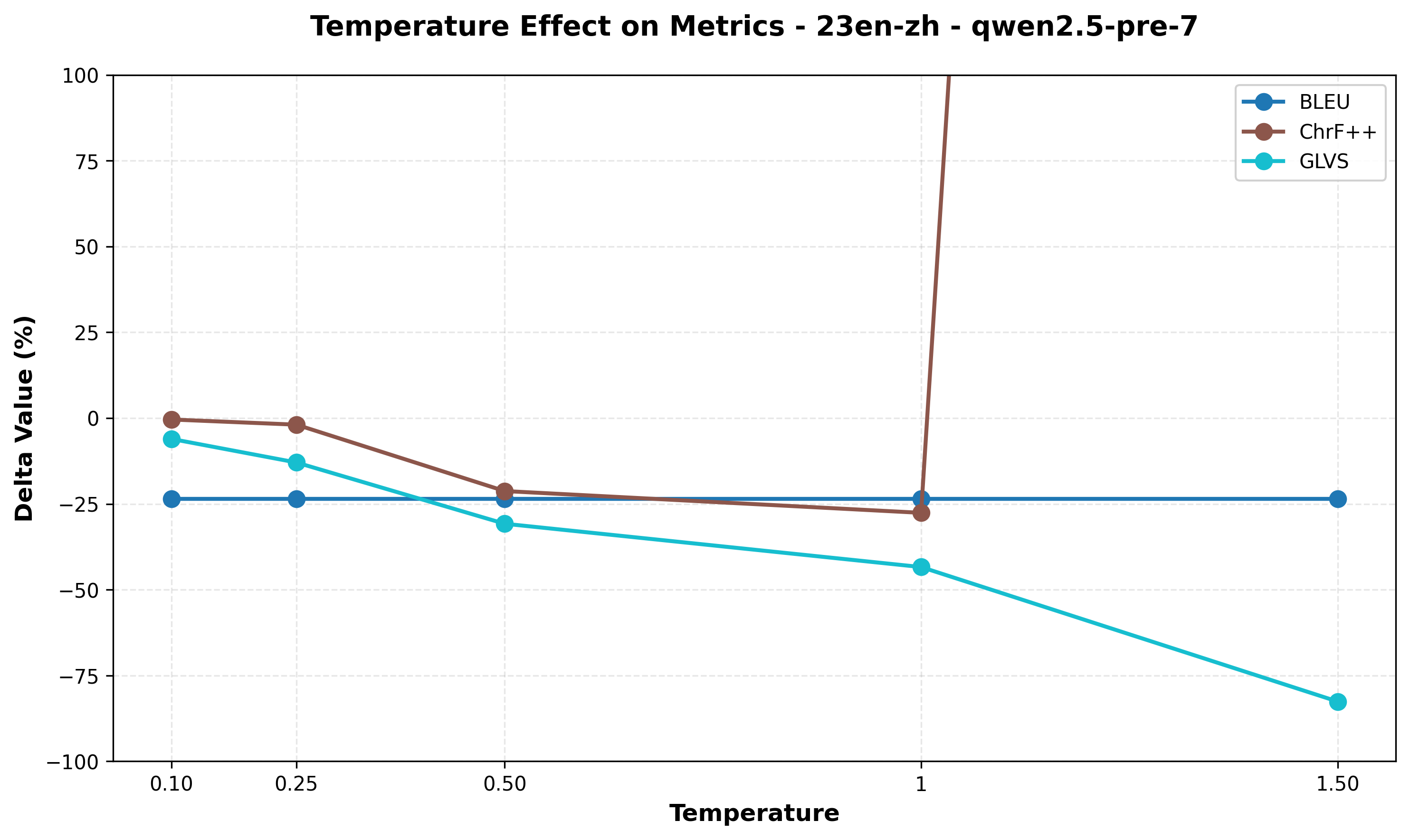}
        \caption{Qwen2.5-Pre-7B (Lexical)}
    \end{subfigure}
    \hfill
    \begin{subfigure}[b]{0.48\linewidth}
        \centering
        \includegraphics[width=\linewidth]{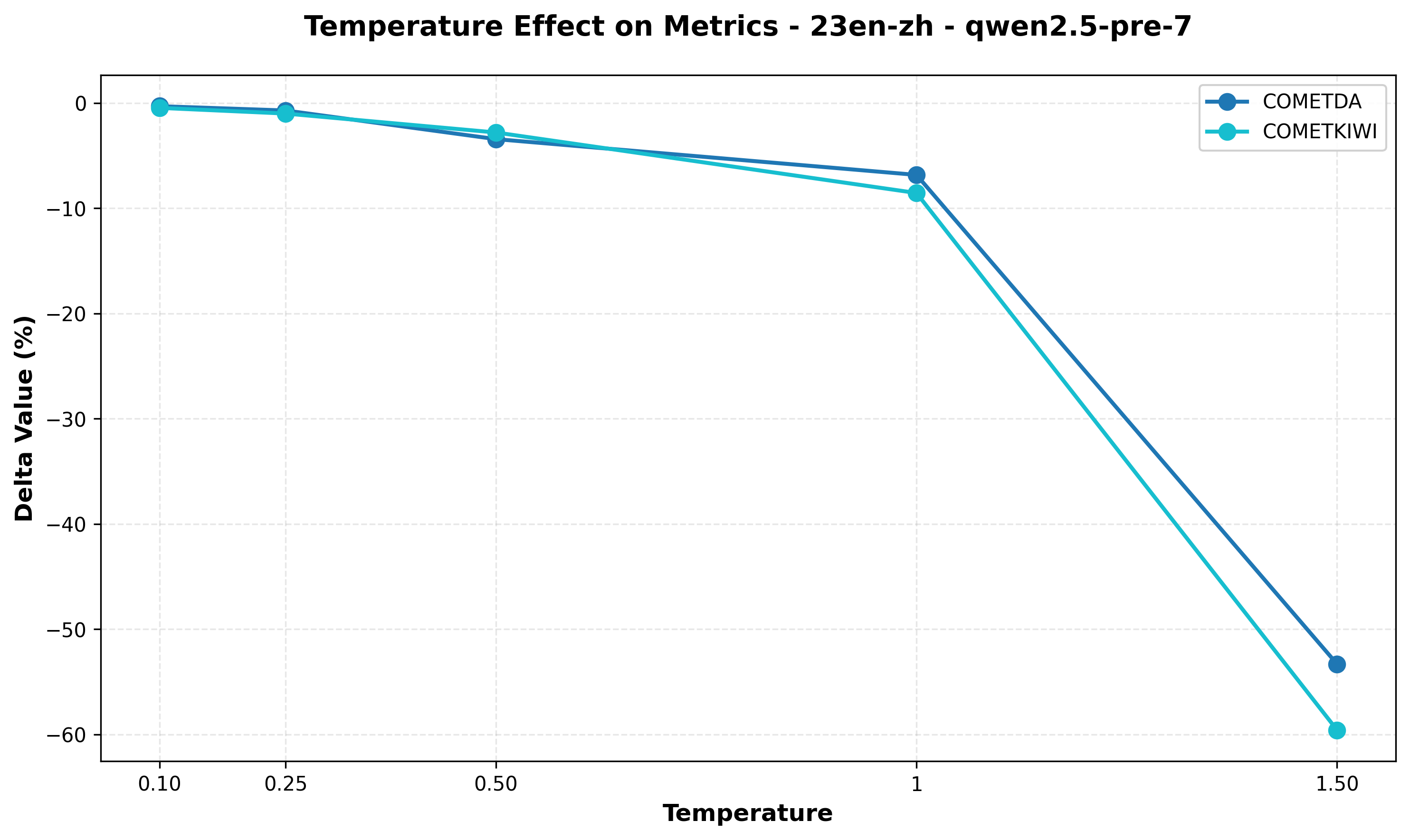}
        \caption{Qwen2.5-Pre-7B (Semantic)}
    \end{subfigure}

    \vspace{0.2cm}

    \begin{subfigure}[b]{0.48\linewidth}
        \centering
        \includegraphics[width=\linewidth]{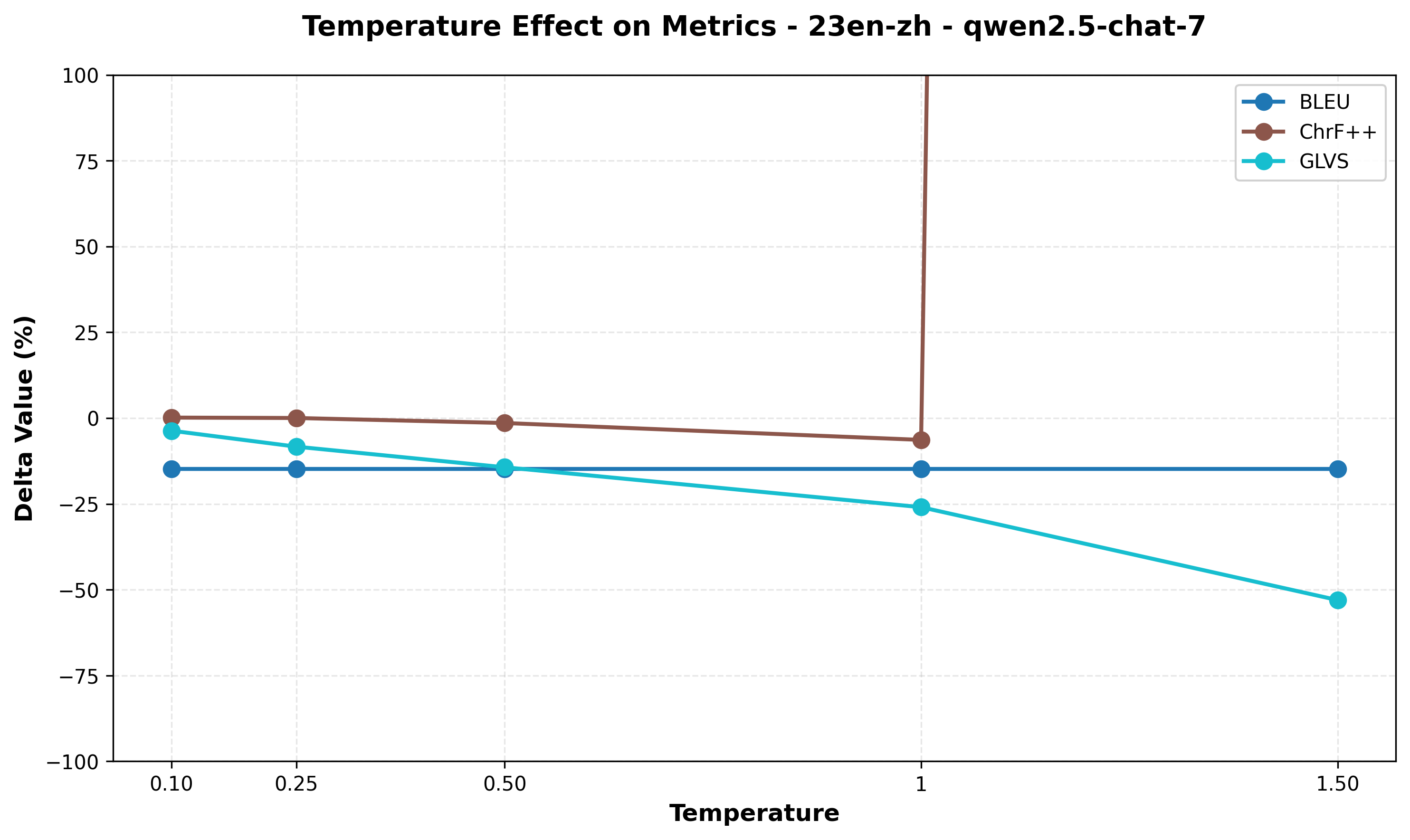}
        \caption{Qwen2.5-Chat-7B (Lexical)}
    \end{subfigure}
    \hfill
    \begin{subfigure}[b]{0.48\linewidth}
        \centering
        \includegraphics[width=\linewidth]{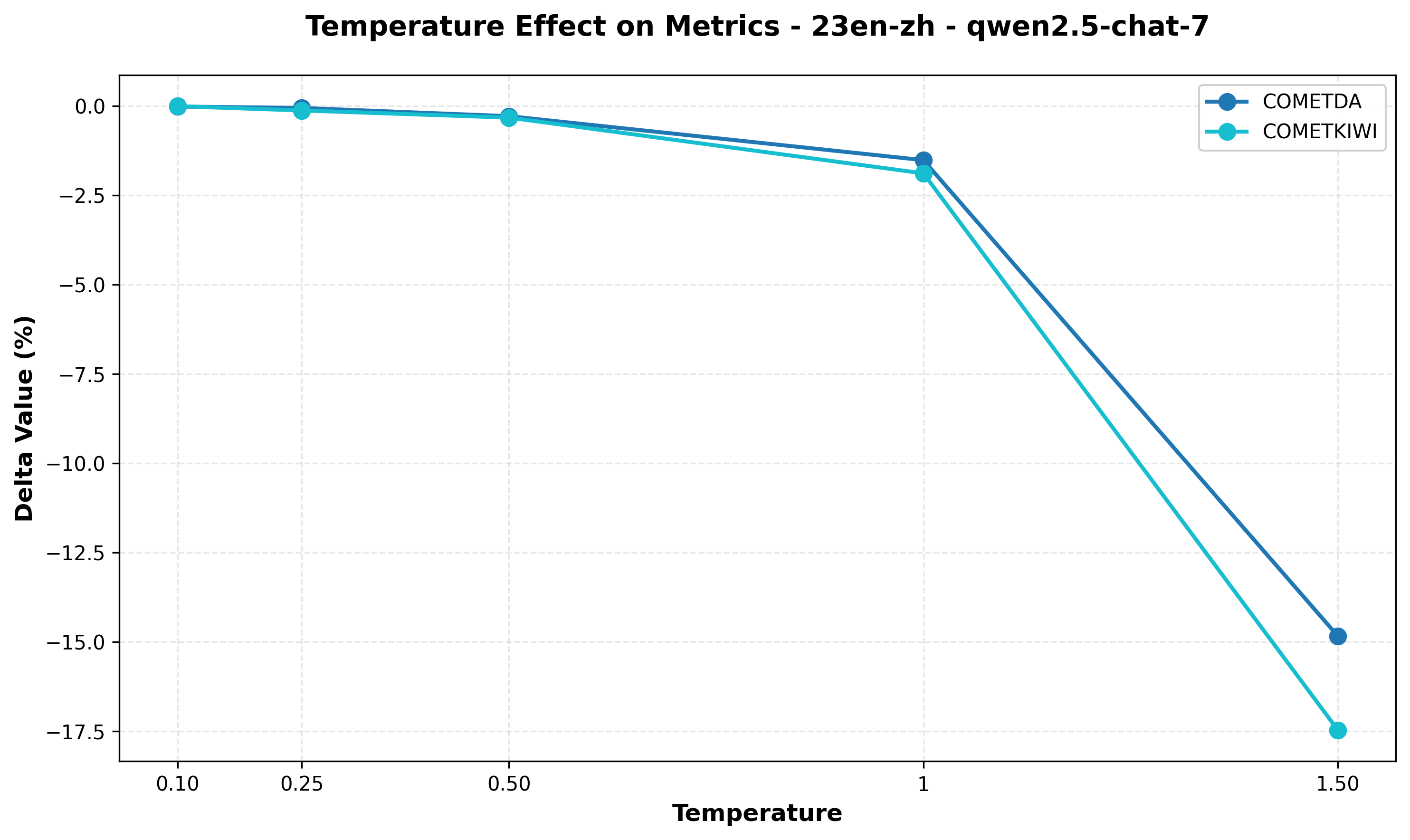}
        \caption{Qwen2.5-Chat-7B (Semantic)}
    \end{subfigure}
    
    \vspace{0.2cm}

    \begin{subfigure}[b]{0.48\linewidth}
        \centering
        \includegraphics[width=\linewidth]{lexical_23en-zh_qwen3-chat-8_temperature_lines.png}
        \caption{Qwen3-Chat-8B (Lexical)}
    \end{subfigure}
    \hfill
    \begin{subfigure}[b]{0.48\linewidth}
        \centering
        \includegraphics[width=\linewidth]{semantic_23en-zh_qwen3-chat-8_temperature_lines.png}
        \caption{Qwen3-Chat-8B (Semantic)}
    \end{subfigure}

    \caption{Temperature effect on the Qwen family. Comparing Pre-trained vs. Chat variants, Qwen models show consistent sensitivity to sampling temperature across both lexical (left) and semantic (right) metrics.}
    \label{fig:temp_qwen_family}
\end{figure*}

\paragraph{Semantic Equivalence vs. Temperature.}
As illustrated in \textbf{Figures~\ref{fig:temp_llama_family} and \ref{fig:temp_qwen_family}}, there is a general downward trend in semantic metric scores as temperature increases, indicating that higher randomness often degrades translation fidelity. However, the optimal temperature is not always the greedy setting ($T=0$). For instance, in specific datasets with \texttt{llama2-chat-7}, non-zero temperatures yield marginal improvements, suggesting that ND-MT can occasionally surpass deterministic baselines (D-MT) when tuned correctly.

\paragraph{Lexical Diversity and Metric Sensitivity.}
Regarding lexical analysis, we observe a distinct divergence between metrics. While BLEU~\cite{bleu} scores remain nearly invariant across the temperature range, GLVS scores exhibit significant volatility (see \textbf{Figure~\ref{fig:temp_llama_family}(a) and (c)}). This demonstrates that BLEU fails to capture the subtle variations in lexical selection introduced by sampling. The shift in GLVS at higher temperatures suggests that models drift toward generating more "natural" language content rather than adhering strictly to the source fidelity, a nuance that standard lexical metrics overlook. These findings highlight the necessity of multi-dimensional evaluation for ND-MT systems.

\begin{table}[H]
\centering
\caption{Correlation Results of Semantic-based Metrics on WMT23 EN-ZH for Five SOTA ND-MT Systems.}
\label{tab:metrics_corr_sample_semantic}
\scriptsize
\setlength{\tabcolsep}{2pt}
\begin{tabular}{@{}l@{\hspace{2pt}}cccc@{}}
\toprule
Strategy& BERT & BLEURT & COMETDA & KIWI \\
\midrule
\multicolumn{5}{@{}l}{\textit{Kendall's $\tau$ / p-value}} \\
\midrule
Min (Worst) & .58/.00 & .64/.00 & .74/.00 & .77/.00 \\
Max (Best) & .57/.00 & .59/.00 & .67/.00 & .70/.00 \\
Mean & .63/.00 & .66/.00 & .73/.00 & .77/.00 \\
Random & .63/.00 & .67/.00 & .73/.00 & .77/.00 \\
Std & -.57/.00 & -.64/.00 & -.71/.00 & -.76/.00 \\
\midrule
\multicolumn{5}{@{}l}{\textit{Spearman's $\rho$ / p-value}} \\
\midrule
Min (Worst) & .72/.00 & .81/.00 & .87/.00 & .89/.00 \\
Max (Best) & .74/.00 & .78/.00 & .83/.00 & .85/.00 \\
Mean & .79/.00 & .84/.00 & .87/.00 & .89/.00 \\
Random & .79/.00 & .85/.00 & .87/.00 & .89/.00 \\
Std & -.69/.00 & -.79/.00 & -.87/.00 & -.89/.00 \\
\bottomrule
\end{tabular}
\begin{tablenotes}
\scriptsize
\item KIWI=COMETKIWI.
\end{tablenotes}
\end{table}
\begin{table*}[t!]
\centering
\caption{Correlation Analysis of Lexical Metrics across Sampling Sizes (20, 50) for Five SOTA ND-MT Systems. The ``Worst Case'' strategy consistently predicts system ranking. Note that for accuracy metrics (BLEU, etc.), \textbf{Min} is the worst case. For the error metric \textbf{TER}, \textbf{Max} is the worst case.}
\label{tab:metrics_corr_lexical}
\small 
\setlength{\tabcolsep}{4pt} 
\begin{tabular}{@{}cl cccccccc@{}}
\toprule
\multirow{2}{*}{\textbf{Size}} & \multirow{2}{*}{\textbf{Strategy}} & \multicolumn{2}{c}{\textbf{BLEU} ($\uparrow$)} & \multicolumn{2}{c}{\textbf{GLVS} ($\uparrow$)} & \multicolumn{2}{c}{\textbf{METEOR} ($\uparrow$)} & \multicolumn{2}{c}{\textbf{ROUGE-1} ($\uparrow$)} \\
\cmidrule(lr){3-4} \cmidrule(lr){5-6} \cmidrule(lr){7-8} \cmidrule(lr){9-10}
& & $\rho$ / $\tau$ & p-val & $\rho$ / $\tau$ & p-val & $\rho$ / $\tau$ & p-val & $\rho$ / $\tau$ & p-val \\
\midrule
\multirow{5}{*}{20} 
& Max (Best) & .70 / .60 & .19 & 1.0 / 1.0 & .00 & 1.0 / 1.0 & .00 & 1.0 / 1.0 & .00 \\
& Mean       & .90 / .80 & .04 & .90 / .80 & .04 & 1.0 / 1.0 & .00 & 1.0 / 1.0 & .00 \\
& \textbf{Min (Worst)} & \textbf{1.0 / 1.0} & .00 & \textbf{1.0 / 1.0} & .00 & \textbf{1.0 / 1.0} & .00 & \textbf{1.0 / 1.0} & .00 \\
& Random     & .90 / .80 & .04 & .90 / .80 & .04 & 1.0 / 1.0 & .00 & 1.0 / 1.0 & .00 \\
& Std        & .70 / .60 & .19 & .90 / .80 & .04 & 1.0 / 1.0 & .00 & .82 / .74 & .09 \\
\midrule
\multirow{5}{*}{50}
& Max (Best) & .70 / .60 & .19 & .90 / .80 & .04 & .90 / .80 & .04 & 1.0 / 1.0 & .00 \\
& Mean       & .90 / .80 & .04 & .90 / .80 & .04 & 1.0 / 1.0 & .00 & 1.0 / 1.0 & .00 \\
& \textbf{Min (Worst)} & \textbf{1.0 / 1.0} & .00 & \textbf{1.0 / 1.0} & .00 & \textbf{1.0 / 1.0} & .00 & \textbf{1.0 / 1.0} & .00 \\
& Random     & .90 / .80 & .04 & .90 / .80 & .04 & 1.0 / 1.0 & .00 & 1.0 / 1.0 & .00 \\
& Std        & .70 / .60 & .19 & 1.0 / 1.0 & .00 & .97 / .95 & .00 & .82 / .74 & .09 \\
\midrule[\heavyrulewidth]
\multirow{2}{*}{\textbf{Size}} & \multirow{2}{*}{\textbf{Strategy}} & \multicolumn{2}{c}{\textbf{ROUGE-2} ($\uparrow$)} & \multicolumn{2}{c}{\textbf{ROUGE-L} ($\uparrow$)} & \multicolumn{2}{c}{\textbf{TER} ($\downarrow$)} & \multicolumn{2}{c}{\textbf{ChrF++} ($\uparrow$)} \\
\cmidrule(lr){3-4} \cmidrule(lr){5-6} \cmidrule(lr){7-8} \cmidrule(lr){9-10}
& & $\rho$ / $\tau$ & p-val & $\rho$ / $\tau$ & p-val & $\rho$ / $\tau$ & p-val & $\rho$ / $\tau$ & p-val \\
\midrule
\multirow{5}{*}{20}
& Max        & .90 / .80 & .04 & .90 / .80 & .04 & \textbf{.90 / .80 (Worst)} & .04 & .90 / .80 & .04 \\
& Mean       & 1.0 / 1.0 & .00 & 1.0 / 1.0 & .00 & .90 / .80 & .04 & 1.0 / 1.0 & .00 \\
& Min        & 1.0 / 1.0 & .00 & 1.0 / 1.0 & .00 & \textit{.40 / .40 (Best)} & .50 & 1.0 / 1.0 & .00 \\
& Random     & 1.0 / 1.0 & .00 & 1.0 / 1.0 & .00 & .90 / .80 & .04 & 1.0 / 1.0 & .00 \\
& Std        & .92 / .88 & .03 & .82 / .74 & .09 & .90 / .80 & .04 & 1.0 / 1.0 & .00 \\
\midrule
\multirow{5}{*}{50}
& Max        & .80 / .60 & .10 & 1.0 / 1.0 & .00 & .90 / .80 & .04 & .90 / .80 & .04 \\
& Mean       & 1.0 / 1.0 & .00 & 1.0 / 1.0 & .00 & .90 / .80 & .04 & 1.0 / 1.0 & .00 \\
& Min        & 1.0 / 1.0 & .00 & 1.0 / 1.0 & .00 & .90 / .80 & .04 & 1.0 / 1.0 & .00 \\
& Random     & 1.0 / 1.0 & .00 & 1.0 / 1.0 & .00 & .80 / .60 & .10 & 1.0 / 1.0 & .00 \\
& Std        & .92 / .89 & .03 & .76 / .67 & .13 & .90 / .80 & .04 & .90 / .80 & .04 \\
\bottomrule
\end{tabular}
\begin{tablenotes}
\scriptsize
\item $\rho$ = Spearman; $\tau$ = Kendall. Values are coefficient / p-value. For TER, Max is Worst Case, Min is Best Case.
\end{tablenotes}
\end{table*}

\begin{table*}[t]
\centering
\caption{Correlation Results of Semantic-based Metrics on WMT23 EN-ZH Comparing Sample Sizes $N=20$ and $N=50$.}
\label{tab:corr_semantic_combined}
\small
\setlength{\tabcolsep}{5pt}
\begin{tabular}{@{}lcccc@{}}
\toprule
\multirow{2}{*}{Strategy} & \multicolumn{2}{c}{COMETDA} & \multicolumn{2}{c}{KIWI} \\
\cmidrule(lr){2-3} \cmidrule(l){4-5}
 & $N=20$ & $N=50$ & $N=20$ & $N=50$ \\
\midrule
\multicolumn{5}{@{}l}{\textit{Kendall's $\tau$ / p-value}} \\
\midrule
Min (Worst) & \textcolor{red}{\textbf{1.0}}/.02 & \textcolor{red}{\textbf{1.0}}/.02 & \textcolor{red}{\textbf{1.0}}/.02 & \textcolor{red}{\textbf{1.0}}/.02 \\
Max (Best) & .40/.48 & .80/.08 & 1.00/.02 & .80/.08 \\
Mean & 1.00/.02 & 1.00/.02 & 1.00/.02 & 1.00/.02 \\
Random & 1.00/.02 & 1.00/.02 & 1.00/.02 & 1.00/.02 \\
Std & .84/.05 & .95/.02 & 1.00/.02 & .80/.08 \\
\midrule
\multicolumn{5}{@{}l}{\textit{Spearman's $\rho$ / p-value}} \\
\midrule
Min (Worst) & \textcolor{red}{\textbf{1.0}}/.00 & \textcolor{red}{\textbf{1.0}}/.00 & \textcolor{red}{\textbf{1.0}}/.00 & \textcolor{red}{\textbf{1.0}}/.00 \\
Max (Best) & .60/.28 & .90/.04 & 1.00/.00 & .90/.04 \\
Mean & 1.00/.00 & 1.00/.00 & 1.00/.00 & 1.00/.00 \\
Random & 1.00/.00 & 1.00/.00 & 1.00/.00 & 1.00/.00 \\
Std & .89/.04 & .97/.00 & 1.00/.00 & .90/.04 \\
\bottomrule
\end{tabular}
\begin{tablenotes}
\scriptsize
\item KIWI=COMETKIWI. N denotes sample size.
\end{tablenotes}
\end{table*}

\section{Case Study: The Impact of Decoding Temperature}
\label{sec:case_study_temp}

\begin{table}[H] 
  \centering
  \small
  \caption{Quantitative analysis of character ratio vs. temperature.}
  \begin{tabular}{lr}
    \toprule
    Temperature ($T$) & Mean Character Ratio $\downarrow$ \\
    \midrule
    0.10 & 0.9541 \\
    0.25 & 0.9467 \\
    0.50 (Default) & 0.7588 \\
    1.00 & 1.2258 \\
    1.50 & \textcolor{red}{45.6444} \\
    \bottomrule
  \end{tabular}
  \label{tab:temp_mean_metric_llama2_enzh}

  \vspace{1.5em} 
\end{table}

\begin{table}[H]
  \centering
  \small
  \begin{CJK*}{UTF8}{gbsn} 
  \caption{Qualitative examples showing temperature-induced collapse.}
  \label{gene_break}
  \begin{tabular}{cp{0.75\linewidth}}
    \toprule
    $T$ & Candidate Translation (Excerpt) \\
    \midrule
    0.10 & 黑客版的星際戰士遊戲在崩潰時是因為它在函數表上請求了一個函數在結束的端點。 \\
    0.25 & 黑客版的星 Wars 專項熱點在 Crash 時是因為它在 vtable 的結尾請求函數。 \\
    0.50 & 《黑客版的杰尼卡特之夜在崩溃中因为它在执行一个函数...》 \\
    1.00 & 「差了！我在《绝地当amentalion》中发现了一个大错误！」 \\
    1.50 & \textit{The hacked up version of Jedi Knight was crashing because it was calling a function off the end of a vtable. China Catholicstructor$))\{\}$......} \\
    \bottomrule
  \end{tabular}
  \end{CJK*} 
  
\end{table}
\paragraph{Generation Length Analysis}
To investigate the empirical effects of decoding temperature on generation stability, we perform a case study using Llama-2-7b-chat on the WMT23 En$\rightarrow$Zh translation task. We employ the character-count ratio (length of translation divided by length of source) as a proxy metric for model stability. 

As shown in Table~\ref{tab:temp_mean_metric_llama2_enzh}, the character ratio remains relatively stable at lower temperatures but undergoes a dramatic "explosion" as $T$ approaches 1.5. This signifies a total loss of structural alignment and linguistic coherence.

\paragraph{Qualitative Analysis of Generation Collapse}
We illustrate this phenomenon using a specific source sentence from the dataset: 
\textit{``The hacked up version of Jedi Knight was crashing because it was calling a function off the end of a vtable.''}

As shown in Table~\ref{gene_break}. At $T=1.5$, the ND-MT system fails to maintain translation constraints. The output diverges into a mixture of garbled multilingual fragments and echoing English source tokens. This observation validates our characterization of modern MT systems as \textbf{temperature-constrained}: while they offer valuable lexical diversity at moderate settings, they lack the inherent robustness to remain semantic equivalence.

\section{The Buckets Effect in ND-MT}
\label{Buckets_Effect}

We introduce the ``Buckets Effect'' hypothesis to characterize ND-MT performance: just as a bucket's capacity is determined by its shortest plank, an ND-MT system's overall reliability is best approximated by its worst-case output. We validate this hypothesis by analyzing the correlation between different aggregation strategies (Min, Max, Mean, Random, Std) and the final system ranking across the five state-of-the-art ND-MT systems.\looseness=-1

\paragraph{Worst-Case Performance Determines Ranking.}
Table~\ref{tab:metrics_corr_lexical} presents the correlation analysis across lexical metrics. The results strongly validate the Buckets Effect. For accuracy metrics (BLEU, GLVS, METEOR, ROUGE), the \textbf{Min} strategy (representing the lowest/worst score) consistently achieves near-perfect correlations ($\rho \approx 1.0$) with the true system ranking. In contrast, the \textbf{Max} strategy (best-case sample) often shows weaker correlations (e.g., $\rho=0.70$ for BLEU), suggesting that a model's ``lucky'' best generations are poor predictors of its overall capability. This trend extends to semantic-based metrics (COMETDA, KIWI) shown in Table~\ref{tab:corr_semantic_combined}. The \textbf{Min} strategy maintains perfect correlations ($\tau=1.00, \rho=1.00$) across both sample sizes ($N=20$ and $N=50$), indicating that the lower bound of generation quality is a robust indicator of system ranking. Conversely, the \textbf{Max} strategy proves unstable, dropping as low as $\tau=0.40$ and $\rho=0.60$ for COMETDA at $N=20$, though it improves with larger sampling sizes. Furthermore, the strong correlation of standard deviation (\textbf{Std}) ($\rho \ge 0.89$) reinforces that system consistency—specifically the ability to minimize variance and avoid quality collapse—is more indicative of model superiority than peak performance.

\paragraph{The Deceptive Nature of Best-Case TER.}
TER presents a unique case because it is an error metric (lower is better). Consequently, the \textbf{Min} strategy represents the \textit{best-case} performance (lowest error), while the \textbf{Max} strategy represents the \textit{worst-case} (highest error). As shown in Table~\ref{tab:metrics_corr_lexical} (Size 20), TER exhibits extreme divergence: its best-case performance (\texttt{Min}) loses predictive power ($\rho=0.40$), whereas its worst-case performance (\texttt{Max}) remains highly predictive ($\rho=0.90$). This confirms that the Buckets Effect holds universally: regardless of the metric's direction, the worst-case output is the true determinant of system quality.

\section{Other Supporting Figures}
\begin{figure*}[!t]
    \centering
    \begin{subfigure}{\textwidth}
        \centering
        \includegraphics[scale=0.5]{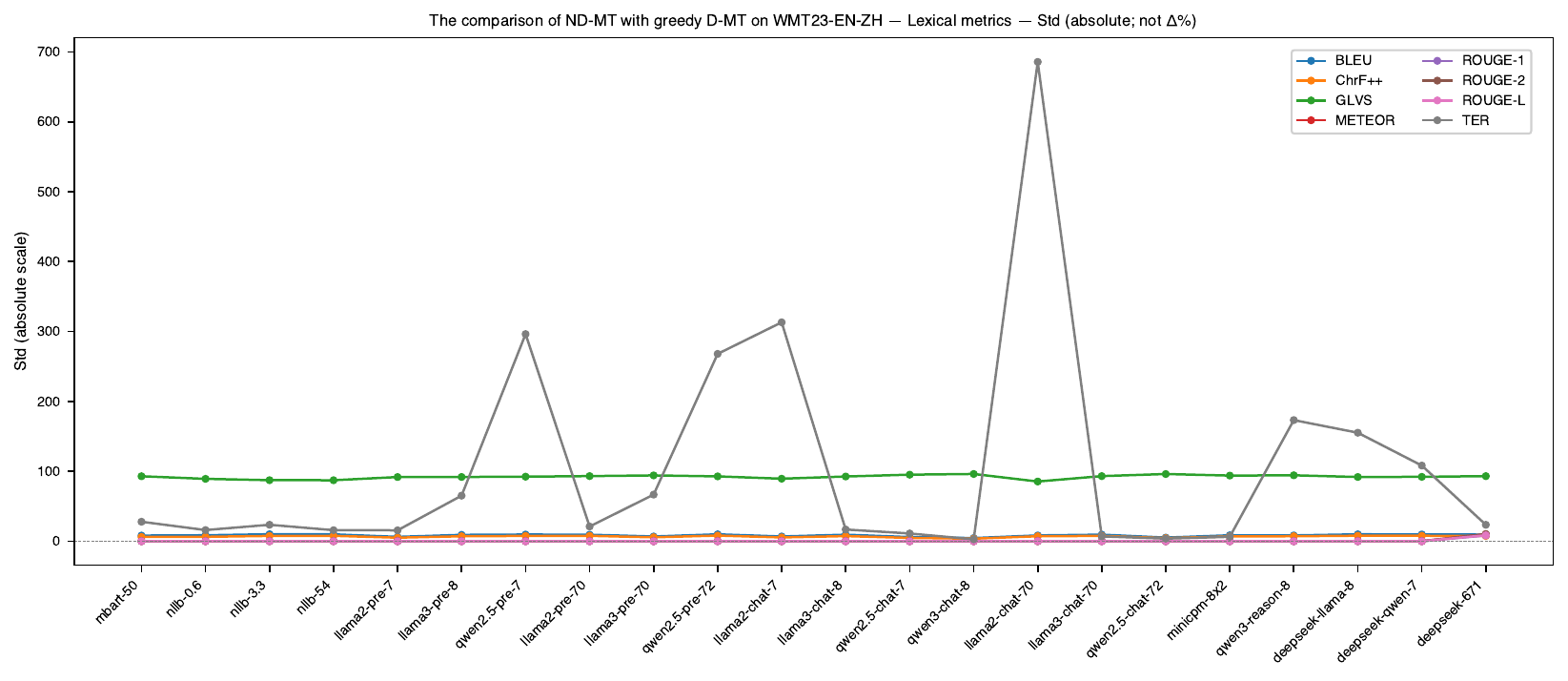}
        \caption{Std Delta Results of WMT23 En$\rightarrow$Zh on Lexical Metrics.}
        \label{fig:delta_lexical_std_en-zh}
    \end{subfigure}
    \begin{subfigure}{\textwidth}
        \centering
        \includegraphics[scale=0.5]{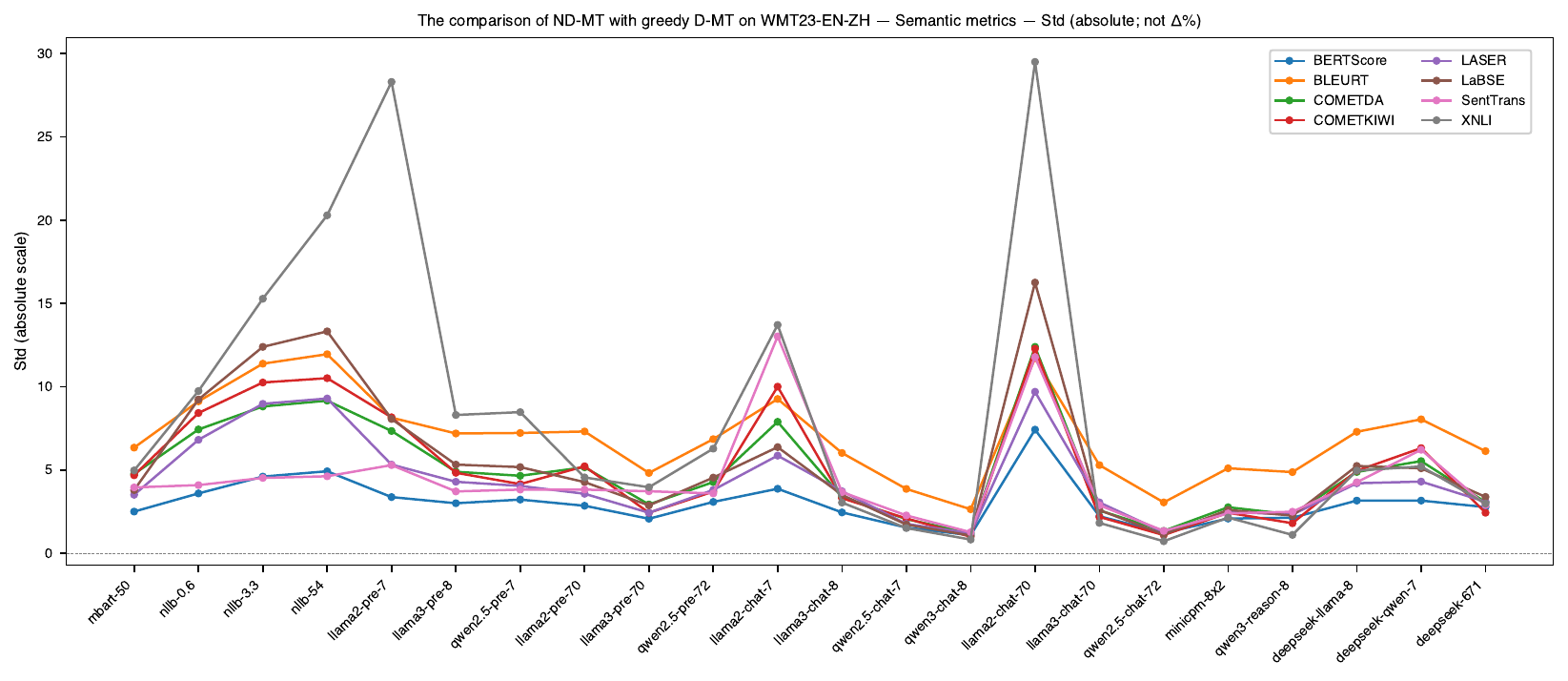}
        \caption{Std Delta Results of WMT23 En$\rightarrow$Zh on Semantic Metrics.}
        \label{fig:delta_semantic_std_en-zh}
    \end{subfigure}
    
    \caption{Std Delta results for both lexical and semantic metrics on WMT23 En$\rightarrow$Zh ($T=0.5$, 10 candidates). Delta results are calculated relative to greedy decoding on identical data and models. Thresholds of $-5$ and $-10$ (dotted line) are included to indicate levels of significance.}
    \label{fig:delta_all_en-zh}
\end{figure*}

\begin{figure*}[!t]
    \centering
    \begin{subfigure}{\textwidth}
        \centering
        \includegraphics[scale=0.5]{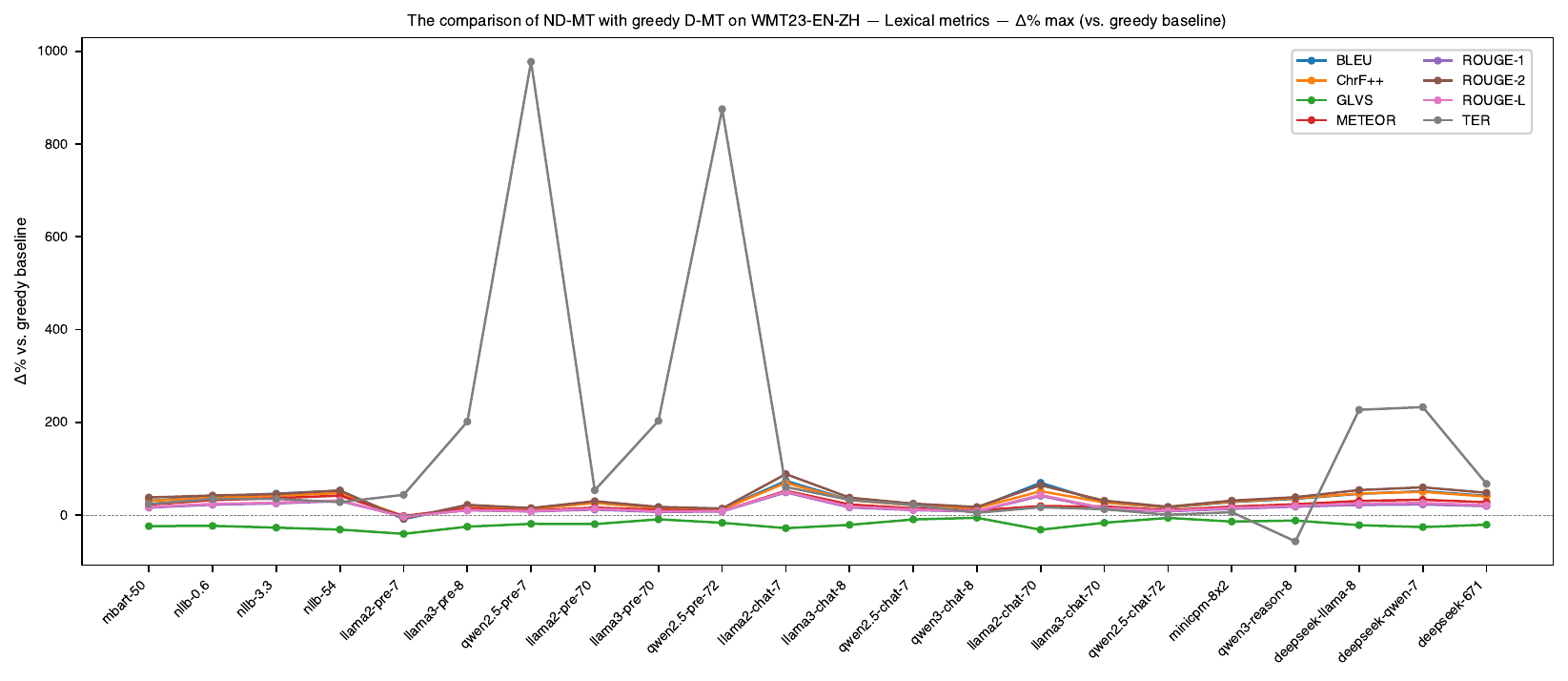}
        \caption{Max delta on lexical metrics (WMT23 En$\rightarrow$Zh).}
        \label{fig:delta_lexical_max_en-zh}
    \end{subfigure}
    \begin{subfigure}{\textwidth}
        \centering
        \includegraphics[scale=0.5]{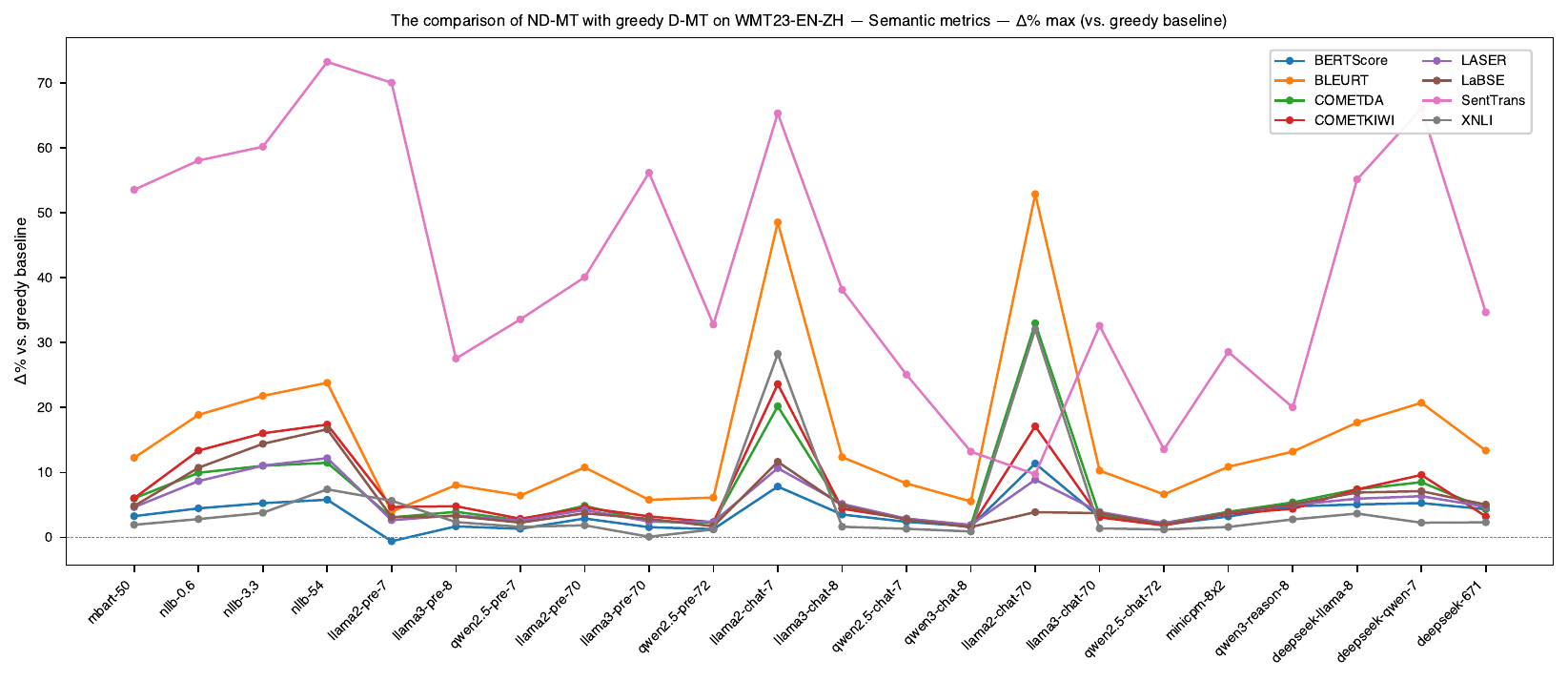}
        \caption{Max delta on semantic metrics (WMT23 En$\rightarrow$Zh).}
        \label{fig:delta_semantic_max_en-zh}
    \end{subfigure}

    \caption{Max delta values on WMT23 En$\rightarrow$Zh for lexical and semantic metrics ($T{=}0.5$, 10 candidates). Deltas are computed relative to greedy decoding on identical data and models.}
    \label{fig:delta_max_en-zh}
\end{figure*}

\begin{figure*}[!t]
    \centering
    \begin{subfigure}{\textwidth}
        \centering
        \includegraphics[scale=0.4]{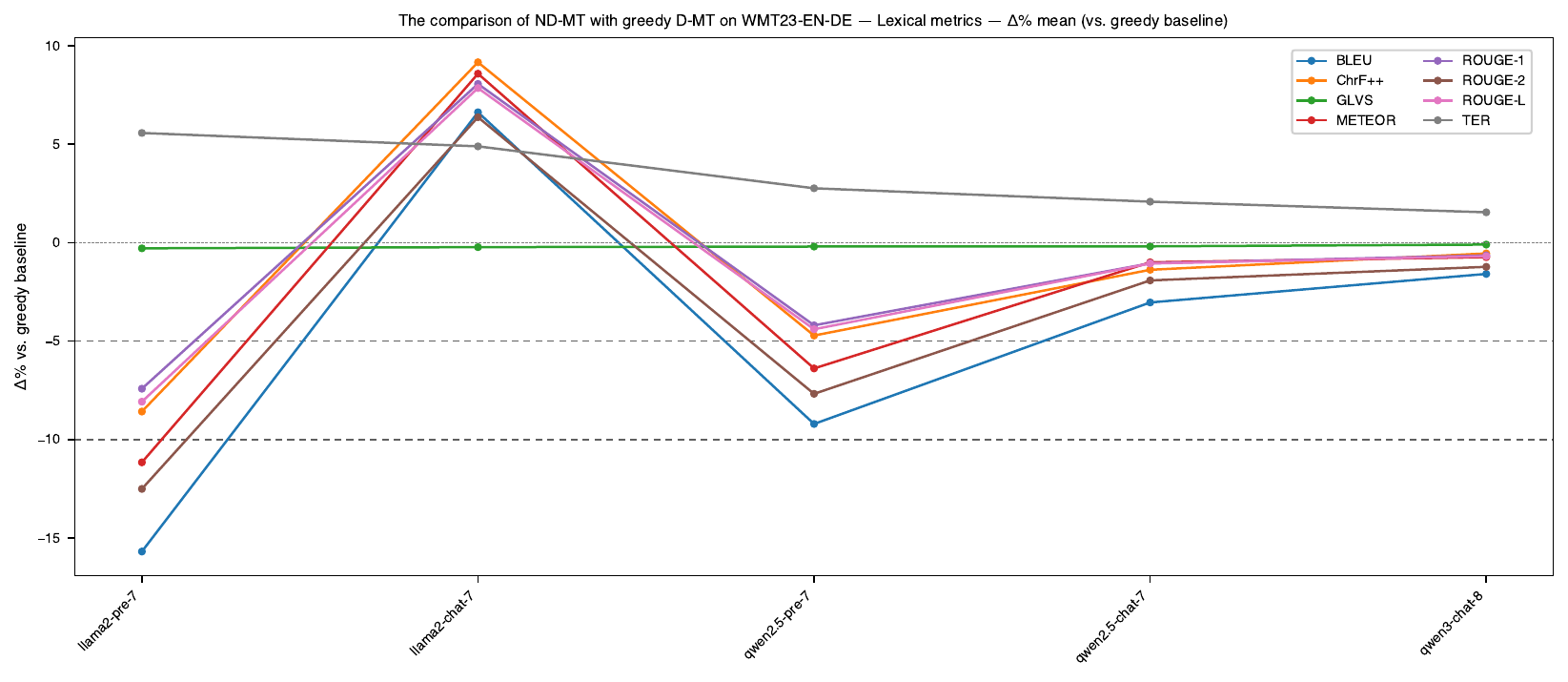}
        \caption{Lexical metrics---delta mean (WMT23 En$\rightarrow$De).}
        \label{fig:delta_lexical_mean_en-de}
    \end{subfigure}
    
    \begin{subfigure}{\textwidth}
        \centering
        \includegraphics[scale=0.4]{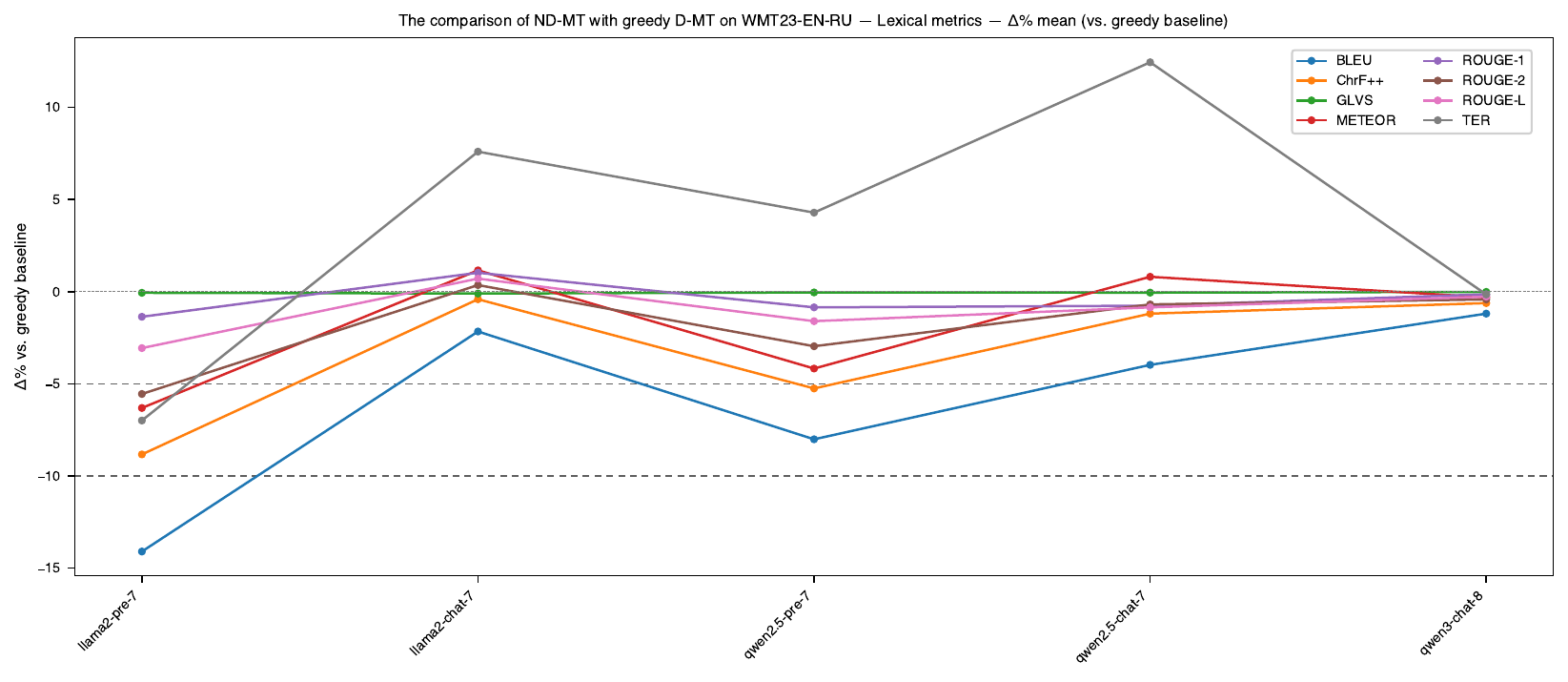}
        \caption{Lexical metrics---delta mean (WMT23 En$\rightarrow$Ru).}
        \label{fig:delta_lexical_mean_en-ru}
    \end{subfigure}

    \vspace{0.5em}

    \begin{subfigure}{\textwidth}
        \centering
        \includegraphics[scale=0.4]{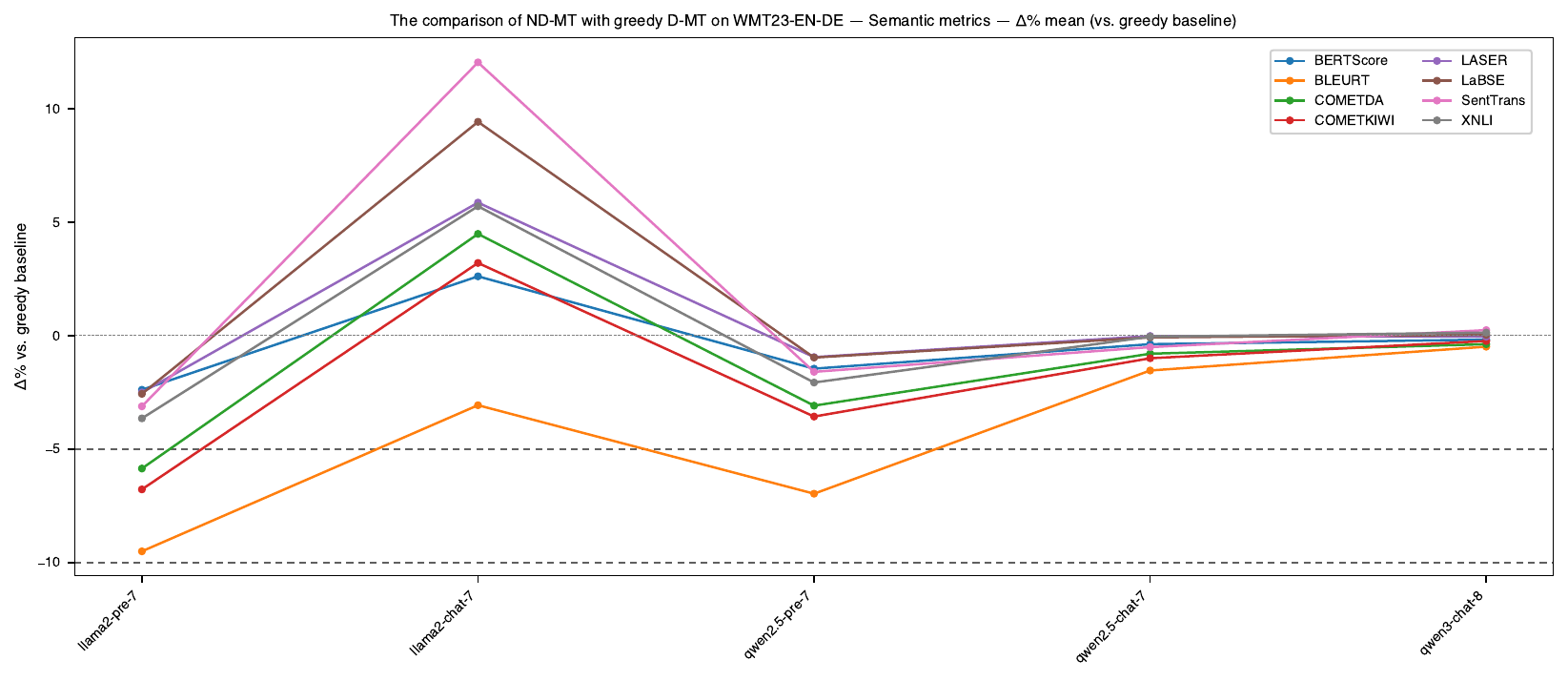}
        \caption{Semantic metrics---delta mean (WMT23 En$\rightarrow$De).}
        \label{fig:delta_semantic_mean_en-de}
    \end{subfigure}
 
    \begin{subfigure}{\textwidth}
        \centering
        \includegraphics[scale=0.4]{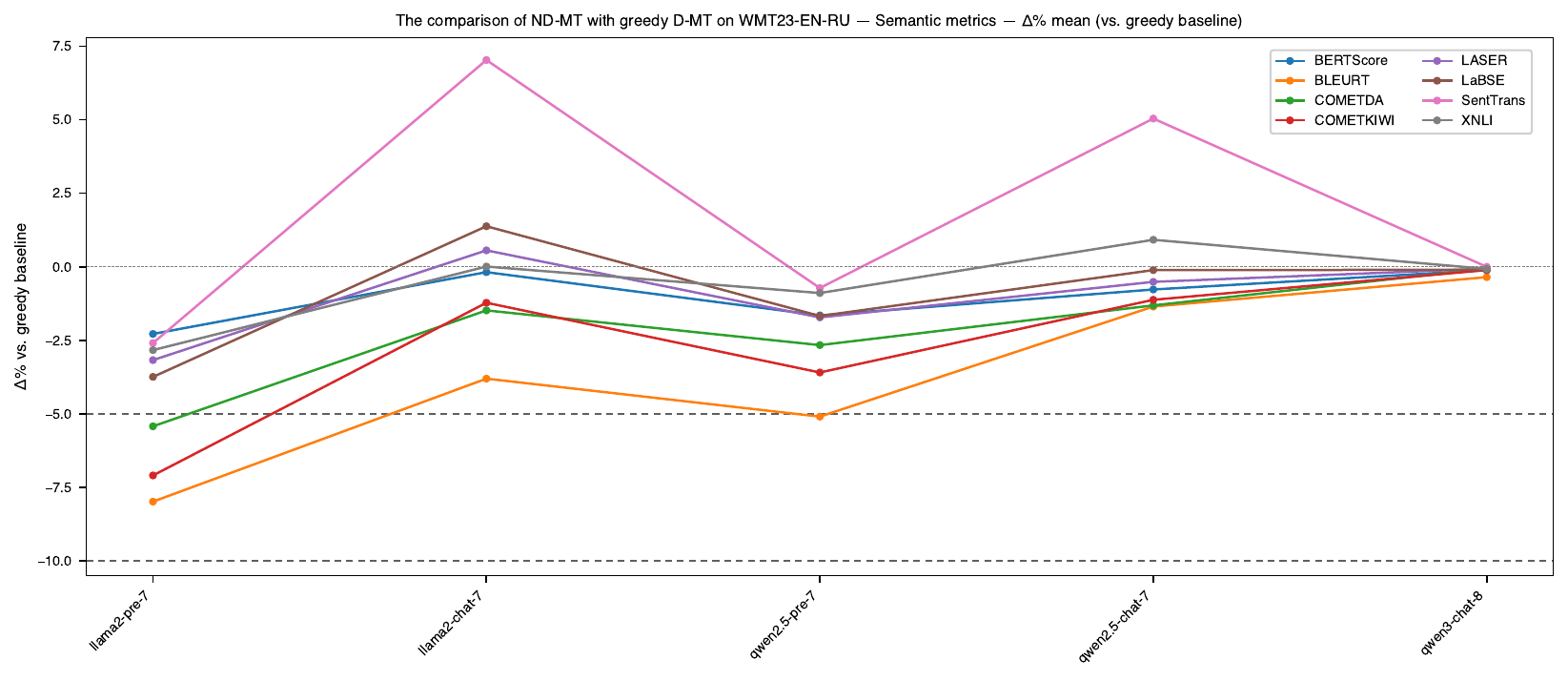}
        \caption{Semantic metrics---delta std (WMT23 En$\rightarrow$Ru).}
        \label{fig:delta_semantic_std_en-ru}
    \end{subfigure}

    \caption{Delta statistics on WMT23 En$\rightarrow$De and En$\rightarrow$Ru for lexical and semantic metrics ($T{=}0.5$, 10 candidates). Deltas are computed relative to greedy decoding on identical data and models.}
    \label{fig:delta_all_en-de-ru}
\end{figure*}

We report the other crucial supporting figures, including the ones to highlight the value of metrics on detecting the generation stability of ND-MT (See Figure~\ref{fig:delta_all_en-zh}); the ones to prove the potential of ND-MT on providing higher quality candidates than D-MT (See Figure~\ref{fig:delta_max_en-zh}); and the ones to show the generality of ND-MT on solving multimodality on diverse language directions (See Figure~\ref{fig:delta_all_en-de-ru}). For more details, please visit our github page.

\section{The Use of AI Assistant}
The authors acknowledge the use of Claude Sonnet 4.5 and Gemini 3 solely for proofreading and polishing the language of this paper (e.g., improving grammar, clarity, and fluency). The writing process incorporated stylistic suggestions under the strict supervision of the authors. All technical ideas, methodology, experiments, analysis, and core content were conceived and produced entirely by the authors, without any AI-based content generation or fabrication.
\end{document}